\documentclass[twoside]{article}

\usepackage[accepted]{aistats2023}

\usepackage{custom}
\newtheorem{assumption}{Assumption}
\newtheorem{theorem}{Theorem}
\newtheorem{lemma}{Lemma}
\usepackage{algorithm}
\usepackage{algorithmic}
\usepackage[caption=false]{subfig}
\newcommand\norm[1]{\lVert#1\rVert}

\usepackage{multirow}
\usepackage{amsmath}
\usepackage{amssymb}
\usepackage{enumitem}
\usepackage{algorithm}

\begin{document}

%

%

\twocolumn[

\aistatstitle{On The Convergence Of Policy Iteration-Based Reinforcement Learning With Monte Carlo Policy Evaluation}

\aistatsauthor{ Anna Winnicki \And R. Srikant }

\aistatsaddress{ University of Illinois  Urbana-Champaign \And  University of Illinois  Urbana-Champaign} ]

\begin{abstract}

A common technique in reinforcement learning is to evaluate the value function from Monte Carlo simulations of a given policy, and use the estimated value function to obtain a new policy which is greedy with respect to the estimated value function. A well-known longstanding open problem in this context is to prove the convergence of such a scheme when the value function of a policy is estimated from data collected from a single sample path obtained from implementing the policy (see page 99 of \cite{sutton2018reinforcement}, page 8 of \cite{tsitsiklis2002convergence}). We present a solution to the open problem by showing that a first-visit version of such a policy iteration scheme indeed converges to the optimal policy provided that the policy improvement step uses lookahead \cite{silver2016mastering, DBLP:journals/corr/MnihBMGLHSK16, silver2017mastering}  rather than a simple greedy policy improvement. We provide results both for the original open problem in the tabular setting and also present extensions to the function approximation setting, where we show that the policy resulting from the algorithm performs close to the optimal policy within a function approximation error.
\end{abstract}

\section{INTRODUCTION}

In many applications of reinforcement learning, the underlying probability transition matrix is known but the size of the state space is large so that one uses approximate dynamic programming methods to obtain the optimal control policy. Examples of such applications include game-playing RL agents for playing games such as Chess and Go. Abstracting away the details, in essence what AlphaZero does is the following \cite{silver2017mastering}: it evaluates the current policy using a Monte Carlo rollout and obtains a new policy using the estimate of the value function of the old policy by using lookahead. We note that AlphaZero collects and uses Monte Carlo returns for all states in each rollout \cite{silver2017mastering}. Thus, effectively the algorithm performs policy iteration using Monte Carlo estimates of the value function. If one ignores the Monte Carlo aspect of policy evaluation but is interested in the tree search of aspects of rollout and lookahead, there are several recent works which quantify the impact of the depth of rollout and lookahead on the performance of algorithm \cite{efroni2019combine, efroni2018, annaor}. However, to the best of our knowledge, there is no analysis of Monte Carlo policy evaluation when the estimates of the value function are obtained from trajectories simulated from the policy. To the best of our knowledge, the only analysis of such algorithms assume that, at each iteration, either one estimates the value function starting from every single state of the underlying MDP \cite{tsitsiklis2002convergence} or from a subset of fixed states \cite{annacdc}. In fact, studying Monte Carlo policy evaluation using a single trajectory from each  policy at each step of policy iteration is a known open problem \cite{sutton2018reinforcement, tsitsiklis2002convergence, suttonbarto}. In this paper, we take a significant step in solving this problem: we prove that, with sufficient lookahead, policy iteration and Monte Carlo policy evaluation does indeed converge provided we use sufficient lookahead during the policy improvement step.

\subsection{Main Contributions}
Our paper has two main contributions.

\paragraph{Convergence of Monte Carlo ES} We prove the convergence of Monte Carlo based policy iteration where a single trajectory corresponding to each policy is used at each iteration to generate returns, or empirical sums of costs that estimate the value function, for states visited by the trajectory. A formulation of this algorithm, which is called Monte Carlo with Exploring Starts (Monte Carlo ES), can be found on page 99 of \cite{sutton2018reinforcement}, and its convergence is ``one of the most fundamental open theoretical questions in reinforcement learning'' (page 99 of \cite{sutton2018reinforcement}). See the Appendix for more on the connection of Monte Carlo ES to practice. 
The work of \cite{tsitsiklis2002convergence} partially solves a variant of Monte Carlo ES, but the results assume a setting that is a hybrid of Monte Carlo sampling using a single trajectory and a generative model. Hence, the convergence of Monte Carlo ES, as well as related variants such as the ``every visit'' version \cite{singh1996reinforcement}, remains an open problem. A major objective of this work is to solve the open problem. 

Modern methods that use policy iteration based algorithms with Monte Carlo methods of policy evaluation have achieved spectacular empirical success in problems with very large state spaces \cite{DBLP:journals/corr/MnihBMGLHSK16, silver2017mastering, silver2017shoji} using lookahead policies computed using Monte Carlo Tree Search (MCTS) as opposed to one-step greedy policies. The motivation behind using the lookahead is to significantly speed up the rate of convergence of  the algorithms. The benefits of using MCTS to compute lookahead policies versus one step greedy policies far outweigh the additional computational overhead which is relatively small when the number of next states and actions is small, which is the case in many problems such as chess and Go. One of our main results shows that, with the use of lookahead, Monte Carlo ES converges asymptotically. We also provide finite-sample error bounds for the algorithms. Since the prior statement of the open problem is in the tabular setting, we present the results for that case. We then extend the results to the case where function approximation is use.  Examples of such applications include game-playing RL agents for playing games such as Chess and Go. 
\paragraph{Extension To Linear Function Approximation} Beyond settling the open problem by using lookahead, we also extend the result to the case where one uses feature vectors to approximate the value function.  We show that when the lookahead is sufficiently large, there is convergence to within a function approximation error. We also provide interpretable finite-sample convergence guarantees. 

Then, we show that our techniques can be easily extended to incorporate other algorithms for policy evaluation with feature vectors that have recently been analyzed. For techniques where the mean square error is known such as TD learning with linear function approximation \cite{srikant2019finite, bhandari2018finite}, we show that the approximation error is approximately equal to the mean square error corresponding to the policy evaluation method with feature vectors. Analogously to the previous extension, we show that when the number of steps of TD learning is very large, the error primarily depends on the function approximation error due to the feature vectors. 
When feature vectors are used, recent approximate policy iteration algorithms have a bound on the error in approximate policy iteration as a function of the discount factor $\alpha$ of $1/(1-\alpha)^2$ (see \cite{bertsekas2019reinforcement, annaor,parr}). When $\alpha$ is very close to 1, which is often the case in practice, reducing the bound by a factor of $1/(1-\alpha)$ significantly improves the performance of the algorithms. In our algorithms, our bounds are approximately of the order $1/(1-\alpha^{H-1})(1-\alpha),$ where $H$ is the amount of lookahead.
\subsection{Related Works}
The connection between Monte Carlo methods and control methods based on policy iteration has been widely studied \cite{sutton2018reinforcement, singh1996reinforcement}. The work of \cite{tsitsiklis2002convergence} studies Monte Carlo sampling with infinitely long trajectories beginning at all states or all states with regular frequencies to perform policy iteration. The works of \cite{Yuanlong, liu} study a similar method in the setting of the stochastic shortest path problem. A related result has been obtained in \cite{wang2022on, joseph} under the strong assumption that for the optimal policy the transient states of the resulting Markov chain form an acyclic graph. 

Monte Carlo methods with infinitely long trajectories and fixed starting states to perform approximate policy iteration with feature vectors for function approximation was studied in \cite{annacdc}. 
The use of rollouts to produce an $m$-step return, where $m$ is the partial evaluation parameter in the Monte Carlo simulation, as opposed to infinitely long trajectories, has been studied in \cite{Puterman1978ModifiedPI, tsitsiklisvanroy, efroni2019combine, annaor} (see Section \ref{section 2} for definitions of return and rollout). More broadly speaking, these methods form a subset of approximate policy iteration algorithms that have been extensively studied; see  \cite{bertsekastsitsiklis,bertsekas2019reinforcement,Puterman1978ModifiedPI} for results on dynamic programming and \cite{scherrer,efroni2020online,tomar2020multistep,efroni2018multiple,9407870} for applications to reinforcement learning. 

The work of \cite{efroni2019combine} uses rollouts in the algorithms for policy evaluation along with multiple-step greedy policies, also known as lookahead policies, which have been featured in recent prominent implementations \cite{DBLP:journals/corr/MnihBMGLHSK16, silver2016mastering, silver2017mastering}. The work of \cite{annaor} defines the necesssity of depth of lookahead and amount of return required for approximate policy iteration as a function of the feature vectors and quanities bounds on the asympotic error. Here, we build upon the work of \cite{annaor} and further strengthen the bounds using stochastic approximation as well as expand the setting of the problem to more carefully understand the role lookahead plays on an algorithm that requires only a single trajectory for each policy at each iteration for convergence. The work of \cite{annacdc} provides a partial connection to the work of \cite{annaor} and the present work as it incorporates stochastic approximation but only in a partially generative model setting, similar to the one in \cite{annaor}. See \cite{Bertsekas2011ApproximatePI, bertsekas2019reinforcement} for more on feature vectors in approximate policy iteration. The works of \cite{Bertsekas2011ApproximatePI} and \cite{bertsekas2019reinforcement} also study a variant of policy iteration wherein a greedy policy is evaluated approximately using feature vectors at each iteration.

When the model of the state space is not known, lookahead policies are computed using the Monte Carlo Tree Search (MCTS) algorithm, which has been studied in \cite{ shah2020nonasymptotic, ma2019monte, munosbook, browne, kocisszepesvari,efroni2018multiple,powell2021reinforcement}. For more on the use of tree search in RL algorithms, see \cite{bertsekas2019reinforcement, baxter, veness, lanctot2014monte}. Lookahead also bears much relationship to Model Predictive Control (MPC) \cite{bertsekas2022lessons}. 

Our algorithms involve a general framework which allows for general methods of policy evaluation using feature vectors followed by policy improvement using lookahead. See \cite{srikant2019finite, bhandari2018finite} for more on policy evaluation with feature vectors.

\section{BACKGROUND ON REINFORCEMENT LEARNING}  \label{section 2}
We consider a finite-state finite-action Markov decision process (MDP). The state space is $\scriptS$ and has cardinality $|\scriptS|$. The action space is $\scriptA$ and has size $|\scriptA|$. 
The probability of transitioning to state $j$ from state $i$ when action $u$ is taken is $P_{ij}(u)$. The associated cost is $g(i,u).$ We assume $g(i,u) \in[0,1]$ $\forall i,u,$ with probability 1.

Policy $\mu: \scriptS \to \scriptA$ prescribes an action to take at state $i \in \scriptS.$ When a policy $\mu$ is fixed, we denote by $g_{\mu} \in \mathbb{R}^{|\scriptS|}$ the vector of expected costs associated with policy $\mu.$ 
We call $P_\mu$ the probability transition matrix corresponding to the associated Markov chain. At time $k$, we call the state of the Markov chain $x_k.$ Consider policy $\mu.$ The associated value function with discount factor $\alpha \in (0,1)$ is given by $J^{\mu}$ defined as follows:
$$
    J^{\mu}(i) := E[\sum_{k=0}^\infty \alpha^k g(x_k, \mu(x_k))|x_0 = i] \quad \forall i \in \scriptS.$$
Herein, we assume that $\alpha \in (0,1)$ for all discount factors $\alpha.$ It is well known that $J^\mu$ solves the associated Bellman equation:
$$
        J^\mu = g_\mu + \alpha P_\mu J^\mu.$$
The associated Bellman operator, $T_\mu: \scriptS \to \scriptS$, is defined as follows: 
$$
    T_\mu J = g_\mu + \alpha P_\mu J. $$ 
When we apply the Bellman operator to vector $J$, the result is called $T_{\mu}J,$ which has the following property:
$$    \norm{T_{\mu}J-J^{\mu}}_\infty\leq \alpha\norm{J-J^{\mu}}_\infty.
$$ If operator $T_{\mu}$ is applied $m$ times to vector $J \in \mathbb{R}^{|\scriptS|},$ then we say that we have performed an $m$-step rollout of the policy $\mu$ and the result $T^m_\mu J$ of the rollout is called the return. See \cite{annaor} for more on rollout. 

Our objective is to find a policy $\mu$ which minimizes the expected discounted cost:
$$
    E[\sum_{k=0}^\infty \alpha^k g(x_k, \mu(x_k))|x_0 = i] \quad \forall i \in \scriptS. \label{eq:objective unbiased}$$
We call the associated value function $J^*$, or the optimal value function. In other words,
$$
    J^* := \min_\mu J^\mu.$$
In order to find $J^*$ and a corresponding optimal policy, we define the Bellman optimality operator $T$. When there is no ambiguity, we call $T$ the Bellman operator. We define the Bellman operator $T: \mathbb{R}^{|\scriptS|} \to \mathbb{R}^{|\scriptS|}$ as follows:  
$$
TJ = \min_\mu E[ g_\mu + \alpha P_\mu J].$$
Component-wise, we have the following:
$$
TJ(i) = \underset{u}\min \Big [ g(i, u) + \alpha \sum_{j=1}^{|\scriptS|} P_{ij}(u)J(j) \Big ].$$
For any vector $J$, we say that the policy corresponding to $TJ$ is the \textit{greedy} policy. When we apply the Bellman operator $H$ times to vector $J$, we denote the resulting operator, $T^H,$ as the $H$-step ``lookahead'' corresponding to $J$. We call the greedy policy corresponding to $T^H J$ the $H$-step lookahead policy, or the lookahead policy, when $H$ is understood. See \cite{annaor} for more definitions on the lookahead policy. More succinctly, the lookahead policy $\mu$ corresponding to vector $V$ is the following:
\begin{align*}
    \mu \in \argmin_{\mu}T^{\mu}T^{H-1} V.
\end{align*}

We have that every time the Bellman operator is applied to vector $J$ to obtain $TJ,$
$$    \norm{TJ-J^*}_\infty\leq \alpha\norm{J-J^*}_\infty.
$$
Thus, applying $T$ to obtain $TJ$ gives a better estimate of the value function than $J$, and hence, better lookahead policies than greedy policies.

The Bellman equations state that $J^*$ is a solution to
$$J^* = TJ^*.$$
 It is well known that every greedy policy with respect to the optimal value function $J^*$ is optimal and vice versa \cite{bertsekastsitsiklis}. 

\section{CONVERGENCE OF FIRST-VISIT TRAJECTORY-BASED POLICY ITERATION} \label{section 4}
The convergence of the Monte Carlo with Exploring Starts (Monte Carlo ES) algorithm (page 99 of \cite{sutton2018reinforcement}) is unknown and is     ``one of the most fundamental open theoretical questions in reinforcement learning'' (page 99 of \cite{sutton2018reinforcement}). 

The episodic algorithm iteratively alternates between policy improvement and evaluation using a single trajectory corresponding to the policy in each episode. For states visited by the trajectory, sums of costs beginning at those states are available and constitute estimates of the value function at the states visited by the trajectory. 
Then, the estimates of the value function at the states visited by the trajectory are used to update components of a vector which stores estimates of the optimal value function of all states for the states visited by the trajectory. Then the next greedy policy is determined from the updated estimate of the optimal value function and the iterative process continues.

\paragraph{Algorithm:} We consider a version of the Monte Carlo ES algorithm similar to the main algorithm in \cite{tsitsiklis2002convergence}, which provides a ``partial solution'' of the open problem. At each iteration, $k$, the algorithm stores an estimate of the optimal value function, $V_k \in \mathbb{R}^{|\scriptS|}$. Using $V_k,$ just as in policy iteration, the algorithm obtains a trajectory corresponding to the lookahead policy (see Section \ref{section 2}) corresponding to $V_k$, $\mu_{k+1}$, where 
\begin{align*}
    \mu_{k+1} = \argmin_{\mu}T^{\mu}T^{H-1} V_k.
\end{align*}We call the set of states visited by the trajectory $\scriptD_k.$ Note that as stated in \cite{annaor, efroni2019combine}, the lookahead policy only needs to be computed for states in $\scriptD_k.$ Additionally, while the computation of $T^{H-1}V_k(i)$ for $i \in \scriptD_k$ may be infeasible, in practice, techniques such as Monte Carlo Tree Search (MCTS) are employed \cite{silver2017mastering}, which are particularly useful when the number of next states and actions is small. 

The trajectory is then used to obtain estimates of $J^{\mu_{k+1}}(i)$ for $i \in \scriptD_k$, which we call $\hat{J}^{\mu_{k+1}}(i)$. In order to obtain $\hat{J}^{\mu_{k+1}}(i)$ for $i \in \scriptD_k$, we perform an $m$-step rollout (see Section \ref{section 2}) with policy $\mu_{k+1}$ by obtaining a discounted sum of $m$ costs beginning at each $i$ for $i \in \scriptD_k.$ The return gives us a noisy version of $T^m_{\mu_{k+1}}T^{H-1}V_k(i)$ for $i \in \scriptD_k.$ We call $w_k(i)$ the unbiased noise that arises, noting that $0\leq w_k(i)\leq \frac{1}{1-\alpha}$. If a state is encountered more than once by the trajectory, we consider the rollout from the first visit to the state. Using $T^m_{\mu_{k+1}}T^{H-1}V_k(i)+w_k(i)$ for $i \in \scriptD_k$, we obtain the next iterate as follows:
\begin{align*}
&V_{k+1}(i) \\&= \begin{cases}(1-\gamma_k) V_k + \gamma_{k}(T^m_{\mu_{k+1}}T^{H-1}V_k+w_k) & i \in \scriptD_k \\
V_k(i) &i \notin \scriptD_k.
\end{cases}
\end{align*} 
where $\gamma_k$ and is assumed to be square summable and sums to infinity is the stepsize or learning rate. 

We can write our iterates as follows:
\begin{align*}
    V_{k+1}(i) &= \mathcal{I}_{i \in \scriptD_k} \Big[ (1-\gamma_k)V_k(i) \\&+ \gamma_k (T_{\mu_{k+1}}^m T^{H-1}V_k(i) + w_k(i))\Big] + \mathcal{I}_{i \notin \scriptD_k}\Big[V_k(i)\Big],
\end{align*} where $\mathcal{I}_{i \in \scriptD_k}$ denotes the indicator function which equals one when state $i$ is visited by the trajectory at iteration $k$ and zero otherwise.




With some algebra, can we rewrite $V_{k+1}$ as follows:
\begin{align}
V_{k+1} = (I- \gamma_k P_{k, \mu_k}) V_k + \gamma_k P_{k, \mu_k} (T_{\mu_{k+1}}^m T^{H-1}V_k+ z_k), \label{eq: -_-}
\end{align} where $I$ denotes the $\scriptS \times \scriptS$ identity matrix, $p_{k, \mu_k}(i)$ is the probability that state $i$ is ever visited by the trajectory under policy $\mu_k$, $P_{k, \mu_k}$ is the diagonal matrix where diagonal entries of the matrix correspond to the values of $p_{k, \mu_k}(i)$ for all $i \in \scriptS,$  and $z_k$ satisfies the same properties as $w_k.$ 

Our algorithm is described in Algorithm \ref{alg:alg 2}.
\begin{algorithm} 
\caption{First-Visit Monte Carlo Policy Evaluation For Policy Iteration}\label{alg:alg 2}
\textbf{Input}: $V_0,m, H.$\\
\begin{algorithmic}[1] 
\STATE Let $k=0$.
\STATE Let $\mu_{k+1}$ be such that $T_{\mu_{k+1}}T^{H-1}V_k = T^H V_k$.\\
\STATE Obtain a trajectory using policy $\mu_{k+1}$ and obtain $T^m_{\mu_{k+1}}T^{H-1}V_k(i)+w_k(i)$ for $i \in \scriptD_k$, where $\scriptD_k$ is the set of states visited by the trajectory and $w_k(i)$ for $i \in \scriptD_k$ is unbiased noise from sampling.
\STATE Obtain $V_{k+1}$ as follows
\begin{align*}
V_{k+1}(i) = \begin{cases}(1-\gamma_k) V_k(i) \\+ \gamma_{k}(T^m_{\mu_{k+1}}T^{H-1}V_k(i)+w_k(i)) & i \in \scriptD_k \\
V_k(i) &i \notin \scriptD_k.
\end{cases}
\end{align*} 
\STATE Set $k \leftarrow k+1.$ Go to 2.
\end{algorithmic}
\end{algorithm}

\paragraph{Remark:}
Note that we need not compute $\mu_{k+1}(i)$ for all states $i\in \scriptS$ at instance $k+1.$ We only need to compute $\mu_{k+1}(i)$ for states encountered in the rollout step of the algorithm.

Note the similarity of our algorithm and the algorithm in \cite{tsitsiklis2002convergence}:
\begin{align*}
V_{k+1} = (1- \gamma_k ) V_k + \gamma_k (J^{\tilde{\mu}_{k+1}}+ w_k),
\end{align*} where $\tilde{\mu}_{k+1}$ denotes the greedy policy with respect to $V_k$ (i.e., $H=1$).

The proof of the main algorithm in \cite{tsitsiklis2002convergence} is similar to the main steps of the proof of modified policy iteration \cite{Puterman1978ModifiedPI} and hinges on showing that 
\begin{align*}
\limsup_{k\to\infty} TV_k - V_k \leq  0.
\end{align*}

To show that $\limsup_{k\to\infty} TV_k - V_k \leq  0,$ the proof relies on the following steps:
\begin{align}
TV_{k+1} 
&\leq T_{\tilde{\mu}_{k+1}}V_{k+1}\nonumber \\
&= T_{\tilde{\mu}_{k+1}}((1-\gamma_k)V_k + \gamma_k (J^{\tilde{\mu}_{k+1}}+w_k))\nonumber \\
&= g_{\tilde{\mu}_{k+1}} + \alpha P_{\tilde{\mu}_{k+1}}((1-\gamma_k)V_k + \gamma_k (J^{\tilde{\mu}_{k+1}}+w_k)) \nonumber\\
&= (1-\gamma_k)(g_{\tilde{\mu}_{k+1}} + \alpha P_{\tilde{\mu}_{k+1}}V_k) \nonumber\\&+ \gamma_k (g_{\tilde{\mu}_{k+1}} + \alpha P_{\tilde{\mu}_{k+1}}(J^{\tilde{\mu}_{k+1}}+w_k))\nonumber\\
&= (1-\gamma_k) (TV_k) + \gamma_k (J^{\tilde{\mu}_{k+1}}+\alpha P_{\tilde{\mu}_{k+1}}w_k),\nonumber
\end{align} where $P_{\tilde{\mu}_{k+1}}$ is the probability transition matrix corresponding to the Markov chain induced by policy $\tilde{\mu}_{k+1}.$
We can then subtract $V_{k+1}$ from both sides and easily obtain the stochastic approximation paradigm that allows us to show that $\limsup_{k\to\infty} TV_k - V_k \leq  0:$
\begin{align*}
&\underbrace{TV_{k+1} - V_{k+1}}_{= X_{k+1}} \\&\leq (1-\gamma_k) (\underbrace{TV_k - V_k}_{=X_k}) + \gamma_k(\underbrace{(\alpha P_{\tilde{\mu}_{k+1}}-I)w_k}_{=: v_k}),
\end{align*}  where $I$ denotes the identity matrix and the noise $v_k$ satisfies similar  properties to $w_k.$
The purpose of showing that $\limsup_{k\to\infty} TV_k - V_k \leq  0$ is that asymptotically, we can use monotonicity to show that $J^* \leq J^{\tilde{\mu}_{k+1}}\leq TV_k.$ Since $J^{\tilde{\mu}_{k+1}}$ is upper and lower bounded by contractions with fixed point $J^*,$ we can use stochastic approximation techniques to obtain convergence of our iterates.  

In our algorithm given by its iterates in \eqref{eq: -_-}, it is clear that we cannot perform the steps of the above used in the proof of \cite{tsitsiklis2002convergence}.

We now give Theorem \ref{theorem 1}, which shows that with sufficiciently large lookahead, the iterates in  equation \eqref{eq: -_-} converge to the optimal value function.
\begin{assumption} \label{assume 1}

\begin{enumerate}[label=(\alph*)]

\item The starting state of the trajectory at each instance is drawn from a fixed distribution, $p$, where $p(i)>0\forall i \in \scriptS.$
\item $\alpha^{H-1}+2(1+\alpha^m )\frac{\alpha^{H-1}}{1-\alpha}<1.$

\item $\sum_{i=0}^\infty \gamma_i = \infty$. Also, $\sum_{i=0}^\infty \gamma_i^2 < \infty.$
\end{enumerate}
\end{assumption}
We make several remarks on our assumptions:
\begin{enumerate}[label=(\alph*)]
    \item The first assusmption is what is denoted as ``exploring starts'' (see \cite{sutton2018reinforcement}), and guarantees for all states to be selected infinitely many times. We note that it is straightforward to extend our results to any initial distribution as long as the probability of visiting any state is lower bounded by a constant. In particular, we do not require a fixed probability distribution for the initial state. 
    \item We assume the lookahead is sufficiently large, see previous sections for more on lookahead.
    \item The stepsizes are square summable and sum to infinity, which allows for noise averaging.
\end{enumerate}

\begin{theorem} \label{theorem 1}
Under Assumption \ref{assume 1}, the iterates of Algorithm \ref{alg:alg 2} given in equation \eqref{eq: -_-} converge to $J^*$, the optimal value function, almost surely.
\end{theorem}
The proof is given in the Appendix. 

\subsection{Proof Idea}

The main idea in the proof is the following. With sufficiently large lookahead, we can show that \begin{align}H(V_k):=T^m_{\mu_{k+1}}T^{H-1}V_k \label{eq: defHcont}\end{align} is a contraction towards $J^*$, and hence we can apply stochastic approximation techniques to obtain convergence of $V_k\to J^*$. We note that in equation \eqref{eq: -_-}, we have written $\mu_{k+1}$ as a function of $V_k$ since it is the lookahead policy with respect to $V_k.$ The matrix $P_{k,\mu_k}$ is a diagonal matrix where each diagonal element indicates if the corresponding state is visited by the trajectory. If the matrix were a constant, one can use the techniques of [Tsitsiklis, 2002], but the key challenge for us is that the matrix is history dependent. The key to our proofs lies in the fact that, with sufficient lookahead, the operator $H(V)$ defined in \eqref{eq: defHcont} is a contraction. For clarify, we can alternatively rewrite $T_{\mu}$ as $T_{\mu(V)}$ when $\mu$ is a lookahead policy corresponding to vector $V$. Note that while the operator $T$ is a contraction, when we consider the operator $T_{\mu(V)},$ $\mu$ depends on $V$ because $\mu$ is the lookahead policy with respect to $V.$ Therefore, it is not obvious if $||T_{\mu(V_1)}^m T^{H-1}V_1-T_{\mu(V_2)}^m T^{H-1}V_2||_\infty$ is smaller than $||V_1-V_2||_\infty.$ 

Our proof hinges on the following key Lemma: 
\begin{lemma}
\begin{align*}
    \norm{J^{\mu_{k+1}}-T^{H-1}V_k}_\infty \leq \frac{\alpha^{H-1}}{1-\alpha}\norm{TV_k - V_k}_\infty,
\end{align*} \label{lemma 1}
\end{lemma} where $J^{\mu_{k+1}}$ is the value function corresponding to policy $\mu_{k+1}$ (see Section \ref{section 2}).
We will prove Lemma \ref{lemma 1} in the Appendix. Using Lemma \ref{lemma 1}, we can show that:
\begin{align*}
    &\norm{T^m_{\mu_{k+1}}T^{H-1}V_k-T^{H-1}V_k}_\infty 
   \\&\leq  (\frac{\alpha^{m+H-1}}{1-\alpha}+\frac{\alpha^{H-1}}{1-\alpha})\norm{TV_k - V_k}_\infty.
\end{align*} 

We can now subtract $J^*$ from both sides of the inequality and use the contraction property of the Bellman operator to get:
\begin{align*}
    &\norm{T^m_{\mu_{k+1}}T^{H-1}V_k-J^*}_\infty 
   \nonumber
  \\ &\leq \Big(\alpha^{H-1}+2(1+\alpha^m )\frac{\alpha^{H-1}}{1-\alpha}\Big)\norm{V_k - J^*}_\infty.
\end{align*} 
\subsection{Novelty of the Proof Technique}
Contrasting with the proof technique of \cite{tsitsiklis2002convergence}, we can see that due to the contraction property that follows from the use of lookahead, we can evade the issues that arise when the proof of \cite{tsitsiklis2002convergence} is extended to include trajectory based updates. Additionally, the contraction based property allows us reduce the asymptotic error that arises from feature vector representation using lookahead, which is the topic of the next section.
\paragraph{Remarks:}
In the special case where the Markov chains induced by all policies are irreducible and infinitely long trajectories are obtained, we recover the results of the main algorithm in \cite{tsitsiklis2002convergence}.

Furthermore, suppose we obtain $T^m_{\mu_{k+1}}T^{H-1}V_k(i)+w_k(i)$ for all $i \in \scriptS,$  for \textit{any} $m$ and $H.$ Then, we can write the following iterative sequence:
\begin{align}
V_{k+1} = (1- \gamma_k ) T^{H-1}V_k + \gamma_k (T_{\mu_{k+1}}^m T^{H-1}V_k+ w_k). \label{eq:&}
\end{align}
We will show in the Appendix that the iterates in \eqref{eq:&} converge to $J^*$ a.s.
Hence, we obtain convergence of a generalized version of the main algorithm in \cite{tsitsiklis2002convergence}.

\section{EXTENSIONS OF FIRST-VISIT SIMULATION-BASED POLICY ITERATION TO LINEAR FUNCTION APPROXIMATION}\label{section 5}

When the sizes of the state and action spaces are very large, we can assign a feature vector $\phi(s)\in \mathbb{R}^d$ to each state $s$ of the state space $\scriptS,$ where $d << |\scriptS|$, and at iteration $k$ obtain an estimate of the value function corresponding to $\mu_{k+1}$, $ \phi(s)^\top \theta^{\mu_{k+1}}$, where  $\theta^{\mu_{k+1}} \in \mathbb{R}^d$ and $\theta^{\mu_{k+1}}$ is estimated from the trajectory corresponding to $\mu_{k+1}$. We define $\Phi$ to be a matrix whose rows are the feature vectors.

In this way, instead of storing vectors $V_k \in \mathbb{R}^{|\scriptS|}$, we can instead update vectors $\theta_k \in \mathbb{R}^d$, where $d<<|\scriptS|.$

When a single trajectory corresponding to the lookahead policy is available, there are many ways to estimate $\theta^{\mu_{k+1}}$. We will formulate an algorithm that allows us to analyze general methods of obtaining $\theta^{\mu_{k+1}}$ and provide convergence guarantees and finite-time bounds for the algorithm.

Our main assumption on the method used to estimate $\theta^{\mu_{k+1}}$ is that there exists known $\kappa$ and $\delta_{app}$ such that 
\begin{align}
    \nonumber&\norm{E[\Phi \theta^{\mu_{k+1}}|\scriptF_k]-T^{H-1}\Phi \theta_k}_\infty 
   \\&\leq   \kappa\norm{  T^{H-1}\Phi \theta_k -J^{\mu_{k+1}}}_\infty+ \delta_{app},\label{eq: this}
\end{align} where $\delta_{app}>0$ and 
$
0<\alpha^{H-1}+ \kappa \frac{2\alpha^{H-1}}{1-\alpha}<1.
$
We will later show how $\kappa$ and $\delta_{app}$ can be obtained for different policy evaluation algorithms. 

We present Algorithm \ref{alg:alg 1} and convergence guarantees of the algorithm for general $\kappa$ and $\delta_{app}$ where $\scriptF_k$ denotes the filtration associated with the noise of the algorithm until instance $k$. Note that similarly to Algorithm \ref{alg:alg 2}, we only need to compute Step 2 of  Algorithm \ref{alg:alg 1} (computation of $\mu_{k+1}$) for states visited by the trajectory. 
\begin{algorithm} 
\caption{Function Approximation Algorithm With Trajectory Based Samples and Lookahead}\label{alg:alg 1}
\textbf{Input}: $\theta_0,m, H$ feature vectors $\{ \phi(i) \}_{i \in \scriptS}, \phi(i) \in \mathbb{R}^d$  .\\
\begin{algorithmic}[1] 
\STATE Let $k=0$.
\STATE Let $\mu_{k+1}$ be such that $T_{\mu_{k+1}}T^{H-1}\Phi \theta_k = T^H \Phi \theta_k$.\\
\STATE Obtain a trajectory using policy $\mu_{k+1}$ and obtain $\theta^{\mu_{k+1}}$ where 
\begin{align*}
    &\norm{E[\Phi \theta^{\mu_{k+1}}|\scriptF_k]-T^{H-1}\Phi \theta_k}_\infty 
    \\&\leq   \kappa\norm{  T^{H-1}\Phi \theta_k -J^{\mu_{k+1}}}_\infty+ \delta_{app}
\end{align*} for some $\delta_{app}>0$ and $\kappa$ such that:
\begin{align*}
0<\alpha^{H-1}+ \kappa \frac{2\alpha^{H-1}}{1-\alpha}<1.
3\end{align*}
\STATE \begin{align}
    \theta_{k+1} = (1-\gamma_k)\theta_k + \gamma_k (\theta^{\mu_{k+1}}). 
\end{align}
\STATE Set $k \leftarrow k+1.$ Go to 2.
\end{algorithmic}
\end{algorithm}
\begin{theorem} \label{theorem 2}
Suppose the following conditions hold:
\begin{itemize}
    \item $\sum_{i=0}^\infty \gamma_i = \infty$,  $\sum_{i=0}^\infty \gamma_i^2 < \infty$ 
    \item there exist $\delta_{app}>0,$ and $\kappa>0$ such that $$0<\alpha^{H-1}+ \kappa \frac{2\alpha^{H-1}}{1-\alpha}<1,$$ and 
    \begin{align*}
    &\norm{E[\Phi \theta^{\mu_{k+1}}|\scriptF_k]-T^{H-1}\Phi \theta_k}_\infty 
   \\&\leq  \kappa\norm{  T^{H-1}\Phi \theta_k -J^{\mu_{k+1}}}_\infty+ \delta_{app},
\end{align*} 
\end{itemize} 
Then, almost surely, the following bound holds for iterates $\theta_k$ of Algorithm \ref{alg:alg 1}:
\begin{align*}
     \limsup_{k\to \infty} \norm{\Phi \theta_k - J^*}_\infty \leq  \frac{\delta_{app}}{1-\alpha^{H-1}-\kappa \frac{2\alpha^{H-1}}{1-\alpha}}
\end{align*}
almost surely and that the policies obtained almost surely have the following property:
\begin{align*}
    \limsup_{k \to \infty} \norm{J^{\mu_k}-J^*}_\infty \leq \frac{2\alpha^{H-1}}{1-\alpha}\Big[\frac{\delta_{app}}{1-\alpha^{H-1}-\kappa \frac{2\alpha^{H-1}}{1-\alpha}}\Big].
\end{align*}
\end{theorem}
The proof of Theorem \ref{theorem 2} can be found in the Appendix.
\subsection{Proof Idea}
Using Lemma \ref{lemma 1}, we will show that our $E[\Phi \theta^{\mu_{k+1}}|\scriptF_k]$ is nearly a contraction with respect to $J^*$. To see this, notice that we can use Lemma \ref{lemma 1} to further upper bound the inequality in \eqref{eq: this} as follows:
\begin{align*}
    &\norm{E[\Phi \theta^{\mu_{k+1}}|\scriptF_k]-T^{H-1}\Phi \theta_k}_\infty 
   \\&\leq  \frac{\kappa\alpha^{H-1}}{1-\alpha}\norm{T\Phi \theta_k - \Phi \theta_k}_\infty+ \delta_{app}.
\end{align*} 

We can now subtract $J^*$ from both sides of the inequality and use the contraction property of the Bellman operator to get:
\begin{align*}
    &\norm{E[\Phi \theta^{\mu_{k+1}}|\scriptF_k]-J^*}_\infty 
   \nonumber
  \\ &\leq  (\alpha^{H-1} +\frac{2\kappa\alpha^{H-1}}{1-\alpha})\norm{\Phi \theta_k - J^*}_\infty+ \delta_{app}.
\end{align*} 

Roughly speaking, the above inequality states that the result of the sampling, $\theta^{\mu_{k+1}},$ contracts towards $J^*$ compared to the previous iterate $\theta_k$. We can then use stochastic approximation techniques to obtain convergence of our iterates.

\subsection{Finite-Time Bounds}

\begin{theorem} \label{theorem 3}
Let $\sigma^2$ be an upper bound on the variance of $(\Phi \theta^{\mu_{k+1}}-E[\Phi \theta^{\mu_{k+1}}|\scriptF_k])$ for all $k.$ Then, we have the following finite-time error bound for Algorithm \ref{alg:alg 1}:
\begin{align*}
    &E[\norm{\Phi \theta_k-J^*}_\infty]  \\&\leq \underbrace{\prod_{i=1}^{k-1} a_{i}\norm{\Phi \theta_0-J^*}_\infty}_{\text{initial condition error}} + \underbrace{\delta_{app}\sum_{j=1}^{k-1} \gamma_j \prod_{\ell=j+1}^{k-1} a_\ell}_{\text{error due to function approximation}}\\&+ \underbrace{\sum_{j=1}^{k-1} \gamma_j(\sigma_{j+1}+\sigma_j) \prod_{\ell=j+1}^{k-1} a_\ell}_{\text{error due to noise}},
\end{align*}
where $a_i := 1-\gamma_i(1-\alpha^{H-1}+\frac{2\kappa\alpha^{H-1}}{1-\alpha})$ and $\sigma_j$ is defined recursively as follows:
\begin{align}
\sigma_{j} = \sigma \sqrt{\sum_{i=1}^j \gamma_i^2\prod_{\ell=i+1}^j (1-\gamma_\ell)^2}.
\label{eq: @*}
\end{align} 
\end{theorem}

\paragraph{Interpretation Of Finite-Time Bounds:}
We will now interpret the terms of the finite-time bounds:
\begin{itemize}
\item Initial condition error: This term goes to 0 as $k\to \infty.$ To see this, notice that $0<1-\alpha^{H-1}+\frac{2\kappa\alpha^{H-1}}{1-\alpha}<1$ due to our assumptions in Theorem \ref{theorem 2}. Thus, since $\sum_{i=0}^\infty \gamma_i = \infty$ and  $\sum_{i=0}^\infty \gamma_i^2 < \infty$, we have that $\prod_{i=1}^{k-1} a_{i} \to 0.$ 
\item Error due to function approximation: $\delta_{app}$ can be interpreted as the error that arises from the use of feature vectors. In our later analysis of $\delta_{app}$ for various algorithms, it can be seen $\delta_{app}$ can be made small if the feature vectors are sufficiently expressive. 
\item Error due to noise: This error term is due to Monte Carlo sampling. Since the discounted infinite horizon reward is bounded by $1/(1-\alpha),$ we have that $(\sigma_{j+1}+\sigma_j)$ goes to zero as $j \to \infty.$ For more discussion, see the Appendix.
\end{itemize}
We now obtain $\kappa$ and $\delta_{app}$ for several methods of computing $\theta^{\mu_{k+1}}.$ 

\subsection{First Visit Monte-Carlo Policy Evaluation With Feature Vectors} \label{subsection 4.3}

We will now go back to Algorithm \ref{alg:alg 2} and directly extend the results to include the use of feature vectors and $\theta_k$ instead of $V_k.$ 

Recall in the previous section that a single trajectory corresponding to the lookahead policy $\mu_{k+1}$ is obtained. We denote the states visited by the trajectory as $\scriptD_k.$ Just as in the previous section, for all states $i \in \scriptD_k,$ we obtain $
T_{\mu_{k+1}}^m T^{H-1} \Phi \theta_k(i) + w_k(i).$ Note that analogously to Section \ref{section 4}, we need only to compute the lookahead for states visited by the trajectory. Additionally, we do not need to compute $\Phi \theta_{k}$ - we only need to compute $\Phi \theta_k(i)=\phi(i)^\top \theta_k$ for states $i$ visited by the trajectory or involved in the tree search. 

Now, instead of updating $V_k(i)$ for $i \in \scriptD_k,$ the case in Section \ref{section 4}, we instead obtain $\theta^{\mu_{k+1}} \in \mathbb{R}^d$, which uses $
T_{\mu_{k+1}}^m T^{H-1} \Phi \theta_k(i) + w_k(i)$ for $i \in \scriptD_k$ to construct an estimate of $\theta^{\mu_{k+1}}.$ 

One way we obtain $\theta^{\mu_{k+1}}$ uses linear least squares to obtain the best fitting $\theta_{k+1}.$
We now compute $\theta^{\mu_{k+1}}$ with linear least squares using the term $\hat{J}^{\mu_{k+1}}$ which we will define in the next paragraph:
\begin{align}
    \theta^{\mu_{k+1}} \nonumber&:= \argmin_\theta \frac{1}{2} \norm{(\scriptP_{1, k}\Phi) \theta - \scriptP_{2, k}\hat{J}^{\mu_{k+1}}}_2^2 \\
    &= (\scriptP_{1,k}\Phi)^+\scriptP_{2,k} \hat{J}^{\mu_{k+1}}. \label{eq: ^.^}
\end{align}  

We now explain the above terms:
\begin{itemize}
\item $\Phi$ is a matrix whose rows are the feature vectors
\item $\scriptP_{2,k}$ is a matrix whose elements are a subset of the elements of $\hat{J}^{\mu_{k+1}}$ corresponding to $\scriptD_k$. Since the values of $\hat{J}^{\mu_{k+1}}(i)$ for $i\notin \scriptD_k$ do not affect $\scriptP_{2, k}\hat{J}^{\mu_{k+1}},$ we can define $\hat{J}^{\mu_{k+1}}$ as follows:
\begin{align*}
\hat{J}^{\mu_{k+1}} := T_{\mu_{k+1}}^m T^{H-1} \Phi \theta_k + w_k,
\end{align*} where $w_k := 0$ for states $i$ $\notin \scriptD_k.$ Notice that since $E[w_k(i)|\scriptF_k]=0$ for $i \in \scriptD_k$  we have that $E[w_k|\scriptF_k]=0.$
\item $\scriptP_{1,k}$ is a matrix of zeros and ones such that rows of $\scriptP_{1,k}\Phi$ correspond to feature vectors associated with states in $\scriptD_k.$ 
\item $(\scriptP_{1,k}\Phi)^+$ is the Moore-Penrose inverse of $\scriptP_{1,k}\Phi.$
\end{itemize}

Using the previous paragraph, we rewrite our iterates in \eqref{eq: ^.^} as follows:
\begin{align*}
    \theta^{\mu_{k+1}} 
    &=  (\scriptP_{1,k}\Phi)^+\scriptP_{2,k} (T_{\mu_{k+1}}^m T^{H-1}\Phi \theta_k+w_k).
\end{align*}  

Our estimate of the value function at iteration $k$ is thus given by:
\begin{align*}
    \Phi\theta^{\mu_{k+1}} 
    &=  \underbrace{\Phi(\scriptP_{1,k}\Phi)^+\scriptP_{2,k} }_{=: \scriptM_k}(T_{\mu_{k+1}}^m T^{H-1}\Phi \theta_k+w_k),
\end{align*}   where $\scriptM_k$ is a projection matrix used to estimate the $J^{\mu_{k+1}}$ from samples of $
\hat{J}^{\mu_{k+1}}(i) = T_{\mu_{k+1}}^m T^{H-1} \Phi \theta_k(i) + w_k(i)$ for $i \in \scriptD_k.$

With $\theta^{\mu_{k+1}}$ defined as the above, we now obtain our corresponding $\delta_{app}$ and $\kappa.$ See Appendix for proofs.
\begin{itemize}
\item $\kappa =  1+\alpha^m \delta_{FV}$ where $\delta_{FV} := \sup_k \norm{\scriptM_k}_\infty.$
\item $\delta_{app}$ 
\begin{align*}=\sup_{k, \mu_k}\norm{E[\scriptM_k( J^{\mu_{k+1}}+ w_k)-(J^{\mu_{k+1}}+w_k)|\scriptF_k]}_\infty,\end{align*} where $w_k$ is a martingale difference sequence, meaning that $E[w_k|\scriptF_k]=0.$
\end{itemize}

How to Interpret Terms In The Error?
\begin{itemize}
\item $\delta_{app}:$ Our error terms in the previous theorems mostly hinge on $\delta_{app}$. Since $\scriptM_k$ is a matrix which uses the feature vectors corresponding to states in $\scriptD_k$ to construct an estimate of $\theta^{\mu_{k+1}}$ based on samples of $\hat{J}^{\mu_{k+1}}(i)$ for states $i \in \scriptD_k$ , it is easy to see that $\delta_{app}$ is a measure of the ability of the feature vectors to approximate the value functions corresponding to the lookahead policies. Hence, with increasingly expressive feature vectors, the error terms in Theorem \ref{theorem 2} go to 0. 
\item $\kappa:$ In the presence of sufficiently large lookahead, $\kappa$ does not drastically alter the results in our theorems. However, we note that typically the quantity $m$ denotes the length of the trajectory starting at each state in $\scriptD_k$, so typically $m$ is very large, and hence $\kappa$ is close to 1. 
\end{itemize}
 
\paragraph{Remarks:}
In order to compute $
T_{\mu_{k+1}}^m T^{H-1} \Phi \theta_k(i) + w_k(i)$ for $i \in \scriptD_k$, we do not need to compute $\Phi \theta_k$; we need only compute $\phi(i)^\top \theta_k$ for states $i \in \scriptD_k$ and states $i$ involved in the computation of the tree search at states visited by the trajectory. Recall from Section \ref{section 4} that we only need to compute the lookahead and $\mu_{k+1}$ for states visited by the trajectory. 

Suppose that $\scriptM_k=I,$ i.e. when we obtain an estimate of $$T_{\mu_{k+1}}^m T^{H-1} V_k(i)$$ for all $i \in \scriptS$. Then, $\Phi \theta_k \to J^*$ a.s. This  matches the result of Theorem \ref{theorem 1}. Additionally, we can see that the error bounds mostly depend on the ability of the representative ability of the feature vectors instead of the sizes of the state and action spaces. 

\subsection{Extension To Gradient Descent}
In order to speed up the rate of convergence of our iterates, we instead take several steps of gradient descent towards 
\begin{align}
    \theta^{\mu_{k+1}} \nonumber&= \argmin_\theta \frac{1}{2} \norm{(\scriptP_{1, k}\Phi) \theta - \scriptP_{2, k}\hat{J}^{\mu_{k+1}}}_2^2 ,
\end{align} where $\scriptP_{1, k}$ and $\scriptP_{2, k}$ are defined in the previous subsection.
In other words, when $\eta$ denotes the number of steps of gradient descent we take, for $\ell = 1, 2, \ldots, \eta.$ We recursively compute the following:
\begin{align}
    \theta_{k+1, \ell} &= \theta_{k+1,\ell-1} - \xi  \nabla_\theta c(\theta;\hat{J}^{\mu_{k+1}})|_{\theta_{k+1,\ell-1}}, \label{eq:iterthetaGD}
\end{align} where
$
    c(\theta;\hat{J}^{\mu_{k+1}}) := \frac{1}{2}\min_\theta \norm{(\scriptP_{1, k}\Phi) \theta - \scriptP_{2, k}\hat{J}^{\mu_{k+1}}}_2^2, 
$ $$0<\xi  < \frac{1}{\sigma_{\scriptP_{1,k}\Phi,\max}} ,$$ and $\sigma_{\scriptP_{1,k}\Phi,\max}$ is the largest singular value squared of $\scriptP_{1,k}\Phi,$  and $\theta_{k+1,0}=0.$
We then set $\theta^{\mu_{k+1}} = \theta_{k+1, \eta}.$ We obtain the following $\kappa$ and $\delta_{app}$: 

\begin{itemize}
\item $\kappa =  1+\alpha^m \delta_{FV},\delta_{FV} :=\sup_{k}\norm{(\scriptP_{1,k}\Phi)^+\scriptP_{2,k}}_\infty.$
\item $\delta_{app}=\sup_{k, \mu_k}\norm{E[\scriptM_k( J^{\mu_{k+1}}+ w_k)-J^{\mu_{k+1}}|\scriptF_k]}_\infty$
\begin{align*}
&+  (1-\xi \sigma_{\scriptP_{1,k}\Phi,\max})^\eta\norm{\Phi}_\infty \norm{V_{k,1}}_\infty  \times \\&\norm{\Sigma_{k,1}^{-1}}_\infty \norm{U_k^\top \scriptP_{2,k}\hat{J}^{\mu_{k+1}}}_\infty,
\end{align*}
where the singular value decomposition of $\scriptP_{1, k}\Phi$ is:
$$
\scriptP_{1, k}\Phi 
= U_k\begin{bmatrix}
\Sigma_{k,1}&
0
\end{bmatrix}\begin{bmatrix}
V_{k,1}^\top \\
V_{k,2}^\top
\end{bmatrix} = U_k \Sigma_{k,1}V_{k,1}^\top.$$
where $U_k$ is a unitary matrix, $\Sigma_{k,1}$ is a rectangular diagonal matrix, and $V_k$ is a unitary matrix.
\end{itemize} 
Our proof of the above is in the Appendix. Ultimately, the purpose of gradient descent is to improve the computational efficiency of the least squares algorithm in Subsection \ref{subsection 4.3}. The results show that as $\eta\to\infty,$ the $\delta_{app}$ of the gradient descent algorithm equals to the $\delta_{app}$ from Subsection \ref{subsection 4.3}. The rate of convergence of $\delta_{app}$ of the gradient descent algorithm towards $\delta_{app}$ of Subsection \ref{subsection 4.3} as a function of the number of steps of gradient descent is exponential. 
\subsection{Other Algorithms Including TD-Learning}

Now consider a general mechanism of obtaining $\theta^{\mu_{k+1}}$ from a sample trajectory. We make the assumption that $\theta^{\mu_{k+1}}$ is bounded (which is always the case for methods with a fixed number of iterations for computing $\theta^{\mu_{k+1}}$). When we have a $\delta$ such that for all $\mu_{k+1}:$
$$
    \norm{E[\Phi \theta^{\mu_{k+1}}]-J^{\mu_{k+1}}}_\infty \leq \delta,
$$ i.e., there exists some $\delta$ which is an upper bound of the error of the method of estimating $J^{\mu_{k+1}}$, we can obtain a corresponding $\delta_{app}$ and $\kappa$ as follows:
\begin{align*}
    \norm{E[\Phi \theta^{\mu_{k+1}}]-T^{H-1}V_k}_\infty  
    &\leq   \norm{  T^{H-1}V_k -J^{\mu_{k+1}}}_\infty
    \\&+\norm{E[\Phi \theta^{\mu_{k+1}}]-J^{\mu_{k+1}}}_\infty \\
    &\leq  \norm{  T^{H-1}V_k -J^{\mu_{k+1}}}_\infty+ \delta.
\end{align*}
Thus, when the mean square error is known, $\kappa= 1$ and $\delta_{app} = \delta.$

Recent studies including \cite{srikant2019finite, bhandari2018finite} have obtained finite-time bounds for TD-learning with linear function approximation. The finite-time bounds in Theorem 3 of \cite{bhandari2018finite} are of the following form: for any $\mu_{k+1}$ where the output of the TD-learning algorithm is $\theta^{\mu_{k+1}}$, we have
$$E[\norm{\Phi{\theta^{\mu_{k+1}}}^*-\Phi \theta^{\mu_{k+1}}}_D^2]\leq \delta_{T,\mu_{k+1}},$$
where $\delta_{T,\mu_{k+1}}$ depends on the number of iterations $T$ of the TD-learning algorithm, $\norm{\cdot}_D$ denotes the weighted 2-norm with weights corresponding to the stationary distribution of $\mu_{k+1}$, and $$\norm{\Phi{\theta^{\mu_{k+1}}}^*-J^{\mu_{k+1}}}_D \leq \frac{1}{\sqrt{1-\alpha^2}}\min_{\theta} \norm{\Phi \theta -J^{\mu_{k+1}}}_D,$$ meaning that ${\theta^{\mu_{k+1}}}^*$ approximates $J^{\mu_{k+1}}.$
From this, it can be shown that:
\begin{align*}
\norm{E[\Phi \theta^{\mu_{k+1}}]-J^{\mu_{k+1}}}_\infty \leq &\sup_{\mu_{k+1}} \norm{\Phi {\theta^{\mu_{k+1}}}^*-J^{\mu_{k+1}}}_\infty \\&+ \frac{\sqrt{\delta_{T,\mu_{k+1}}}}{\pi_{\mu_{k+1},\min}},
\end{align*} where $\pi_{\mu_{k+1},\min}$ denotes the minimum weight of the stationary distribution of $\mu_{k+1},$ thus giving us a $\delta$ as desired. See Appendix for proofs with TD-learning.



\section{CONCLUSION}

We study Monte Carlo methods that estimate the value function corresponding to policies determined in the policy improvement step of Monte Carlo based policy iteration methods. We are concerned with trajectory based updates that involve obtaining estimates of the value function corresponding to the greedy policies from states that are visited by the trajectory. This is noted as an open problem in \cite{sutton2018reinforcement} and \cite{tsitsiklis2002convergence}. We show that when lookahead policies, which are commonly used in practice, are employed, we obtain convergence to the optimal value function.
We further our analysis to include the use of feature vectors and also include analyses of general methods of policy evaluation in feature vector space that are computationally efficient such as TD learning.

\subsubsection*{Acknowledgements}
The authors thank the reviewers for their helpful comments. The research presented here was supported by the following: NSF CCF 22-07547, NSF CNS 21-06801, NSF CCF 1934986, ONR N00014-19-1-2566, and ARO W911NF-19-1-0379.

\bibliography{refs}

\appendix
\onecolumn

\section{PROOF OF CONVERGENCE OF ITERATES IN EQUATION (2) OF SECTION 3}
We write our iterates as follows: $$V_{k+1} = (1-\gamma_k) T^{H-1}V_k + \gamma_k (T_{\mu_{k+1}}^m T^{H-1}V_k + w_k).$$ 

We break the proof up into steps as follows.

\noindent\textit{Step 1:} 
\begin{align*}
    \limsup_{k\to\infty} TV_k - V_k \leq  0.
\end{align*}

\textit{Proof of Step 1:}
We will show that for every $\eps>0$, there exists sufficiently large $k(\eps)$ such that the following holds:
\begin{align}
 (1-\gamma_k) T^{H-1}V_k + \gamma_k T_{\mu_{k+1}}^m T^{H-1}V_k -\eps e \leq  V_{k+1} \leq (1-\gamma_k) T^{H-1}V_k + \gamma_k T_{\mu_{k+1}}^m T^{H-1}V_k   +\eps e ,\label{eq: iter wo y}
\end{align}
where $e$ is the vector of all $1$s.

To do this, we define a sequence of random variables, $Y_k$ as follows:
\begin{align*}
    Y_{k+1} = (1-\gamma_k) Y_k + \gamma_k w_k, Y_0 = 0.
\end{align*}

It is clear that $Y_k \to 0$ a.s. by standard stochastic approximation theory. Then, we subtract $Y_{k+1}$ from both sides of the iterates as follows:
\begin{align*}
V_{k+1}-Y_{k+1} &= (1-\gamma_k) (T^{H-1}V_k-Y_k) + \gamma_k (T_{\mu_{k+1}}^m T^{H-1}V_k).
\end{align*}

Rearranging terms, we have:
\begin{align*}
V_{k+1} &= (1-\gamma_k) (T^{H-1}V_k) + \gamma_k (T_{\mu_{k+1}}^m T^{H-1}V_k)+Y_{k+1} -(1-\gamma_k)Y_k .
\end{align*}

Since $Y_k \to 0 $ a.s., we have that for every $\eps >0,$ there exists $k(\eps)$ such that for all $k>k(\eps)$ we have the right side of the inequality in \eqref{eq: iter wo y}. The left side follows accordingly.

Using the inequality in \eqref{eq: iter wo y}, we have that:
\begin{align*}
    TV_{k+1}&\leq T_{\mu_{k+1}}\Big[(1-\gamma_k) T^{H-1}V_k + \gamma_k T_{\mu_{k+1}}^m T^{H-1}V_k   +\eps e\Big] \\
    &= (1-\gamma_k)T^{H}V_k + \gamma_k T_{\mu_{k+1}}^{m+1}T^{H-1}V_k + \alpha \eps e.
\end{align*}

Furthermore, using the inequality in \eqref{eq: iter wo y},
\begin{align*}
    TV_{k+1}-V_{k+1} &\leq (1-\gamma_k)(T^H V_k - T^{H-1}V_k) + \gamma_k (T_{\mu_{k+1}}^{m+1}T^{H-1}V_k - T_{\mu_{k+1}}^m T^{H-1}V_k) + (1+\alpha)\eps e .\end{align*}

We recursively define $\delta_k$ such that:
\begin{align*}
    TV_k - V_k \leq \delta_k e.
\end{align*}

For $k(\eps)$, we have that:
\begin{align*}
    \delta_{k(\eps)} := \norm{TV_k - V_k}_\infty.
\end{align*}

For $k > k(\eps)$, we define $\delta_k$ as follows:
\begin{align*}
   \delta_k =  \delta_{k-1} (\alpha^{H-1}+\alpha^{m+H-1})+ (1+\alpha)\eps.
\end{align*}

It is clear that $TV_k - V_k\leq \delta_k$ since 
\begin{align*}
   & TV_{k-1}-V_{k-1}\leq \delta_{k-1} e\\
   &\implies T^H V_{k-1}-T^{H-1}V_{k-1}\leq \alpha^{H-1}\delta_{k-1} e\\
   &\implies T_{\mu_k}^m T^H V_{k-1}-T_{\mu_k}^m T^{H-1}V_{k-1}\leq \alpha^{m+H-1}\delta_{k-1} e.
\end{align*}

Thus, 
\begin{align*}
    TV_{k+1}-V_{k+1}&\leq (1-\gamma_k)\alpha^{H-1}\delta_{k-1}e + \gamma_k [\alpha^{m+H-1}\delta_{k-1} e]+ (1+\alpha)\eps e\\
    &\leq (1-\gamma_k)\delta_{k-1}(\alpha^{H-1} e+\alpha^{m+H-1})+ (1+\alpha)\eps e \\
    &= \delta_{k} e.
\end{align*}

Thus, we have that 
\begin{align*}
    \limsup_{k\to\infty} TV_k - V_k \leq \lim_{k\to\infty} \delta_k e.
\end{align*}

We now calculate $\lim_{k \to \infty} \delta_k$ as follows:
\begin{align*}
    \lim_{k \to \infty} \delta_k  = \frac{1+\alpha}{\alpha^{H-1} +\alpha^{m+H-1}} \eps .
\end{align*}

Since $\eps$ can be any value greater than 0, we have that $\lim_{k\to\infty}TV_k - V_k \leq 0.$ 

\noindent\textit{Step 2:}  

For all $\eps,\tilde{\eps}>0$,
\begin{align}
    V_{k+1}
    &\leq T^{H-1}V_k +  \frac{\alpha^{H-1}}{1-\alpha}\tilde{\eps}e+\eps e.
\end{align}

\textit{Proof of Step 2:}

Hence, for any $\tilde{\eps}>0,$ there exists   $k(\tilde{\eps})$ such that for any $k >k(\tilde{\eps}),$ $TV_k - V_k  \leq \tilde{\eps} e.$
 
Thus:
\begin{align*}
    &TV_k - V_k  \leq \tilde{\eps} e \\ \allowdisplaybreaks
    &\implies TV_k \leq V_k + \tilde{\eps} e \\ \allowdisplaybreaks
        &\implies T^H V_k \leq T^{H-1} V_k + \alpha^{H-1}\tilde{\eps}e \\ \allowdisplaybreaks
        &\implies T_{\mu_{k+1}}^m T^{H-1} V_k \leq T^{H-1} V_k + \frac{\alpha^{H-1}}{1-\alpha}\tilde{\eps} e. \\ \allowdisplaybreaks
\end{align*}

Thus, we have for $k>k(\eps)+k(\tilde{\eps})$:
\begin{align*}
    V_{k+1} &\leq (1-\gamma_k)T^{H-1}V_k + \gamma_k (T^{H-1}V_k + \frac{\alpha^{H-1}}{1-\alpha}\tilde{\eps}e)+\eps e \\
    &\leq T^{H-1}V_k + \gamma_k  \frac{\alpha^{H-1}}{1-\alpha}\tilde{\eps}e+\eps e\\
    &\leq T^{H-1}V_k +  \frac{\alpha^{H-1}}{1-\alpha}\tilde{\eps}e+\eps e.
\end{align*}

\noindent\textit{Step 3:}  

For all $\eps,\tilde{\eps}>0$,
\begin{align*}
    V_{k+1}
    &\geq T^{m+H-1}V_k -  \frac{\alpha^{H-1}}{1-\alpha}\tilde{\eps}e-\eps e.
\end{align*}

\textit{Proof of Step 3:}
Furthermore, since $TV_k \leq V_k + \tilde{\eps}$ for all $k>k(\tilde{\eps})$, we have that $$T^{H-1}V_k \geq T^{m+H-1}V_k - \frac{\alpha^{H-1}}{1-\alpha}\tilde{\eps} e.$$

Thus:
\begin{align*}
    V_{k+1} &\geq (1-\gamma_k) (T^{m+H-1}V_k-\frac{\alpha^{H-1}}{1-\alpha}\tilde{\eps} e)  + \gamma_k (T_{\mu_{k+1}}^m T^{H-1}V_k )- \eps e\\
    &\geq (1-\gamma_k) (T^{m+H-1}V_k-\frac{\alpha^{H-1}}{1-\alpha}\tilde{\eps}e) + \gamma_k ( T^{m+H-1}V_k )- \eps e\\
    &= T^{m+H-1}V_k -(1-\gamma_k) \frac{\alpha^{H-1}}{1-\alpha}\tilde{\eps}e-\eps e\\
    &\geq T^{m+H-1}V_k - \frac{\alpha^{H-1}}{1-\alpha}\tilde{\eps}e-\eps e.
\end{align*}

\noindent\textit{Step 4:}  

\begin{align*}
   \norm{V_{k+1}-J^*}_\infty \leq \alpha^{H-1}\norm{V_k - J^*}_\infty +  \frac{\alpha^{H-1}}{1-\alpha}\tilde{\eps}+\eps .
\end{align*}

\textit{Proof of Step 4:}

Putting the above together, we have:
\begin{align*}
    T^{m+H-1}V_k - \frac{\alpha^{H-1}}{1-\alpha}\tilde{\eps} e -\eps e \leq V_{k+1} \leq T^{H-1}V_k +  \frac{\alpha^{H-1}}{1-\alpha}\tilde{\eps}e+\eps e. 
\end{align*}

Subtracting $J^*$ and using the contraction property of the Bellman operator, we have:
\begin{align*}
    -\alpha^{m+H-1}\norm{V_k -J^*}_\infty e - \frac{\alpha^{H-1}}{1-\alpha}\tilde{\eps}e -\eps e
    &\leq T^{m+H-1}V_k -J^* - \frac{\alpha^{H-1}}{1-\alpha}\tilde{\eps} e-\eps e\\&\leq V_{k+1} - J^* \\&\leq T^{H-1}V_k -J^*+  \frac{\alpha^{H-1}}{1-\alpha}\tilde{\eps}e+\eps e \\
    &\leq \alpha^{H-1}\norm{V_k - J^*}_\infty e+ \frac{\alpha^{H-1}}{1-\alpha}\tilde{\eps}e+\eps e. 
\end{align*} 

Thus, 
\begin{align*}
   \norm{V_{k+1}-J^*}_\infty \leq \alpha^{H-1}\norm{V_k - J^*}_\infty +  \frac{\alpha^{H-1}}{1-\alpha}\tilde{\eps}+\eps .
\end{align*}

The above implies that:
\begin{align*}
    \limsup_{k\to\infty} \norm{V_k - J^*}_\infty \leq \frac{\frac{\alpha^{H-1}}{1-\alpha}\tilde{\eps}+\eps }{\alpha^{m+H-1}+\alpha^{H-1}}
\end{align*}

Since the above holds for all $\eps>0$ and all $\tilde{\eps}>0$, we have that: \begin{align*}
   V_k \to J^* a.s.
\end{align*}
\section{PROOF OF LEMMA 1}
The following holds:
\begin{align*}
    &T_{\mu_{k+1}}T^{H-1}V_k - T^H V_k =0 \\
    &\implies T_{\mu_{k+1}}T^{H-1}V_k - T^H V_k + T^{H-1}V_k - T^{H-1}V_k=0 \\
    &\implies T_{\mu_{k+1}}T^{H-1}V_k + \norm{T^H V_k - T^{H-1}V_k}_\infty e - T^{H-1}V_k \geq 0\\
    &\implies T_{\mu_{k+1}}T^{H-1}V_k + \alpha^{H-1}\norm{TV_k - V_k}_\infty e -T^{H-1}V_k \geq 0\\
    &\implies J^{\mu_{k+1}} -T^{H-1}V_k\geq  - \frac{\alpha^{H-1}}{1-\alpha}\norm{TV_k - V_k}_\infty e, 
\end{align*}
where the last line follows from iteratively applying $T_{\mu_{k+1}}$ to both sides and using a telescoping sum and $e$ is the vector of all $1$s. 

We also have:
\begin{align*}
    &T_{\mu_{k+1}}T^{H-1}V_k = T^H V_k \\
    &\implies T_{\mu_{k+1}}T^{H-1}V_k - T^H V_k + T^{H-1}V_k - T^{H-1}V_k = 0\\
    &\implies T_{\mu_{k+1}}T^{H-1}V_k - \norm{T^H V_k - T^{H-1}V_k}_\infty e - T^{H-1}V_k \leq 0\\
    &\implies T_{\mu_{k+1}}T^{H-1}V_k - \alpha^{H-1}\norm{TV_k - V_k}_\infty e -T^{H-1}V_k \leq 0 \\
    &\implies J^{\mu_{k+1}} -T^{H-1}V_k\leq   \frac{\alpha^{H-1}}{1-\alpha}\norm{TV_k - V_k}_\infty e.
\end{align*}
Putting the above two together, we get the following:
\begin{align*}
    \norm{J^{\mu_{k+1}}- T^{H-1}V_k}_\infty \leq \frac{\alpha^{H-1}}{1-\alpha}\norm{TV_k - V_k}_\infty.
\end{align*}
\section{PROOF OF THEOREM 1}
We break the proof of Theorem 1 up into steps. 

\noindent\textit{Step 1:} 
\begin{align*}
    \norm{J^{\mu_{k+1}}-T^{H-1}V_k}_\infty \leq \frac{\alpha^{H-1}}{1-\alpha}\norm{TV_k - V_k}_\infty.
\end{align*}

\textit{Proof of Step 1:}
Step 1 is a restatement of Lemma 1 which is proved in Appendix B.

\noindent\textit{Step 2:} 
\begin{align*}
    &\norm{H(V_k)-T^{H-1}V_k}_\infty 
    \leq  ( \frac{\alpha^{m+H-1}}{1-\alpha}+\frac{\alpha^{H-1}}{1-\alpha})\norm{TV_k - V_k}_\infty
\end{align*}
\textit{Proof of Step 2:}
We have:
\begin{align*}
   & \norm{H(V_k)-T^{H-1}V_k}_\infty - \norm{T^{H-1}V_k-J^{\mu_{k+1}}}_\infty\\&\leq \norm{H(V_k)-T^{H-1}V_k + T^{H-1}V_k-J^{\mu_{k+1}}}_\infty\\&= \norm{H(V_k)-J^{\mu_{k+1}}}_\infty
    \\&\leq \alpha^m \norm{  T^{H-1}V_k -J^{\mu_{k+1}}}_\infty,
\end{align*}

Which implies that 
\begin{align*}
    \norm{H(V_k)- T^{H-1}V_k}_\infty \leq (1+\alpha^m )\norm{  T^{H-1}V_k -J^{\mu_{k+1}}}_\infty .
\end{align*}

Plugging in the results of Step 2, we have:
\begin{align*}
    \norm{H(V_k)- T^{H-1}V_k}_\infty \leq (1+\alpha^m )\frac{\alpha^{H-1}}{1-\alpha}\norm{TV_k - V_k}_\infty .
\end{align*}

\noindent\textit{Step 3:} 
\begin{align*}
  \norm{H(V_k)-J^*}_\infty  \leq  \underbrace{\Big(\alpha^{H-1}+(1+\alpha^m )\frac{\alpha^{H-1}}{1-\alpha}(1+\alpha)\Big)}_{=: \beta}\norm{V_k - J^*}_\infty.
\end{align*}
\textit{Proof of Step 3:}
We have
\begin{align*}
   &T^{H-1}V_k - (1+\alpha^m )\frac{\alpha^{H-1}}{1-\alpha}\norm{TV_k-V_k}_\infty e \leq H(V_k) \nonumber\\&\leq T^{H-1}V_k + (1+\alpha^m )\frac{\alpha^{H-1}}{1-\alpha}\norm{TV_k-V_k}_\infty e\\
   &\implies \norm{H(V_k)-T^{H-1}V_k}_\infty \leq (1+\alpha^m )\frac{\alpha^{H-1}}{1-\alpha}\norm{TV_k-V_k}_\infty  \\
   &\implies \norm{H(V_k)-J^*}_\infty -\norm{T^{H-1}V_k-J^*}_\infty \leq (1+\alpha^m )\frac{\alpha^{H-1}}{1-\alpha}\norm{TV_k-V_k}_\infty  \\ 
   &\implies \norm{H(V_k)-J^*}_\infty  \leq\norm{T^{H-1}V_k-J^*}_\infty+
   (1+\alpha^m )\frac{\alpha^{H-1}}{1-\alpha}\norm{TV_k-V_k}_\infty  \\
      &\implies \norm{H(V_k)-J^*}_\infty  \leq \alpha^{H-1}\norm{V_k-J^*}_\infty
      \nonumber\\&+ (1+\alpha^m )\frac{\alpha^{H-1}}{1-\alpha}(1+\alpha)\norm{V_k - J^*}_\infty  \\
          &\implies \norm{H(V_k)-J^*}_\infty  \leq  \underbrace{\Big(\alpha^{H-1}+(1+\alpha^m )\frac{\alpha^{H-1}}{1-\alpha}(1+\alpha)\Big)}_{=: \beta}\norm{V_k - J^*}_\infty.
\end{align*} Note that above, $e$ is a vector of all $1$s.

\noindent\textit{Step 4:} 
\begin{align*}
    V_k \to J^*.
\end{align*}
\textit{Proof of Step 4:}

So far, we have the following rewrite of our iterates:
\begin{align*}
    V_{k+1}(i)
    &= (1-\gamma_kp_{k,\mu_k}(i))V_k(i) + \gamma_k p_{k,\mu_k}(i)( H(V_k)(i) + z_k(i)),
\end{align*}
where 
\begin{align*}
    \norm{H(V_k)-J^*}_\infty \leq \beta\norm{V_k-J^*}_\infty   .
\end{align*}

We define $\Delta_{k}:= V_{k}-J^*$. Using $\Delta_{k},$ the following holds:
\begin{align*}
   \Delta_{k+1}(i) = (1-\gamma_kp_{k,\mu_k}(i)) \Delta_k(i) + \gamma_k p_{k,\mu_k}(i) (H(V_k)-J^*+z_k)(i). 
\end{align*}

Letting $Y_{k}$ be a sequence defined recursively as follows:
\begin{align*}
    Y_{k+1}(i) = (1-\gamma_k p_{k,\mu_k}(i)) Y_k(i) + \gamma_k p_{k,\mu_k}(i) z_k(i),
\end{align*} where $Y_0 =0.$
Since $w_k$ is bounded for all $k$, $Y_k \to 0$ a.s.

We now define the following sequence $X_k$ as follows:
$X_k := \Delta_k - Y_k.$

Thus, 
\begin{align*}
    X_{k+1}(i) = (1-\gamma_k p_{k,\mu_k}(i))X_k(i) + \gamma_k p_{k,\mu_k}(i) (H(V_k)-J^*)(i).
\end{align*}

Taking absolute values on both sides we have:
\begin{align*}
    |X_{k+1}(i)| &= (1-\gamma_k p_{k,\mu_k}(i))|X_k(i)| + \gamma_k p_{k,\mu_k}(i) |(H(V_k)-J^*)(i)| \\
    &\leq (1-\gamma_k p_{k,\mu_k}(i))\norm{X_k}_\infty + \gamma_k p_{k,\mu_k}(i) \norm{H(V_k)-J^*}_\infty \\
    &\leq (1-\gamma_k p_{k,\mu_k}(i))\norm{X_k}_\infty +\gamma_k p_{k,\mu_k}(i) \beta\norm{V_k-J^*}_\infty  \\
    &\leq (1-\gamma_k p_{k,\mu_k}(i))\norm{X_k}_\infty + \gamma_k p_{k,\mu_k}(i) \beta\norm{\Delta_k}_\infty   \ \\
    &\leq (1-\gamma_k p_{k,\mu_k}(i))\norm{X_k}_\infty + \gamma_k p_{k,\mu_k}(i) \beta\norm{X_k}_\infty +\beta \norm{Y_k}_\infty \\
    &\leq \max_i \Big[ (1-\gamma_k p_{k,\mu_k}(i))\norm{X_k}_\infty + \gamma_k p_{k,\mu_k}(i) \beta\norm{X_k}_\infty +\beta \norm{Y_k}_\infty \Big]. 
\end{align*}

We denote by $\tilde{\gamma}_k$ the $\gamma_k p_{k,\mu_k}(i)$ corresponding to a  maximizing $i$ in the above expression. Thus, 
\begin{align*}
    |X_{k+1}(i)|
    \leq (1-\tilde{\gamma}_k)\norm{X_k}_\infty + \tilde{\gamma}_k \beta\norm{X_k}_\infty +\beta \norm{Y_k}_\infty,
\end{align*}

and since the right hand side of the inequality does not depend on $i$, we have that:
\begin{align*}
   \norm{X_{k+1}}_\infty
    \leq (1-\tilde{\gamma}_k)\norm{X_k}_\infty + \tilde{\gamma}_k \beta\norm{X_k}_\infty +\beta \norm{Y_k}_\infty.
\end{align*}

Since $Y_k \to 0$ a.s., we conclude there must exist for all $\eps>0$ some $k(\eps)$ such that for all $k>k(\eps):$
$$
\norm{Y_k}_\infty \leq \eps. 
$$
So, for $k>k(\eps),$ the following holds:
\begin{align*}
    \norm{X_{k+1}}_\infty &\leq  (1-\tilde{\gamma}_k)\norm{X_k}_\infty + \tilde{\gamma}_k \Big[\beta\norm{X_k}_\infty +\beta \eps \Big].
\end{align*}
Rearranging terms, we have:
\begin{align*}
    &\norm{X_{k+1}}_\infty \nonumber\\&\leq (1-\tilde{\gamma}_k(1-\beta))\norm{X_k}_\infty + \tilde{\gamma}_k(\beta \eps ) \\
    &= (1-\underbrace{\tilde{\gamma}_k(1-\beta)}_{=: \gamma_k'})\norm{X_k}_\infty + \tilde{\gamma}_k (1-\beta)\Big[ \frac{\beta \eps }{1-\beta}\Big]. 
\end{align*}

Now, consider any positive integer $N$. We define a sequence of random variables $\overline{X}_k^N$ for $k \geq N,$ by setting $\overline{X}_N^N= \norm{X_N}_\infty$ and 
\begin{align*}
    \overline{X}_{k+1}^N = (1-\gamma_k')\overline{X}_k^N + \gamma_k' \Big[ \frac{\beta \eps }{1-\beta}\Big]\forall k > N.
\end{align*}

We will carry out a comparison of the sequence $\norm{X_k}_\infty$ with the sequence $\overline{X}_k^N$.
Consider the event that $k(\eps)=N,$ which we denote by $A_N.$ We can use an easy inductive argument to show that for any $N$, for any sample path in $A_N,$ and for all $k\geq N,$ that $\norm{X_k}_\infty \leq \overline{X}_k^N.$
It is evident from the assumptions that  $\sum_{k=0}^\infty \gamma_k' = \infty $ and hence $\overline{X}_k^N \to  \frac{\beta \eps }{1-\beta}e$ as $k\to\infty.$
To see this, observe that when the terms of $\overline{X}_k^N$ are written out, we have:
\begin{align*}
    \overline{X}_k^N = \prod_{\ell=N+1}^k (1-\gamma_\ell') \overline{X}_N^N+ (1-\prod_{\ell=N+1}^k (1-\gamma_\ell')) \frac{\beta \eps }{1-\beta}e
\end{align*} for $k >N.$ Since $\gamma_k'$  sums to infinity, we have that $\lim_{k\to\infty}\prod_{\ell=N}^k (1-\gamma_\ell') = 0,$ hence $\overline{X}_k^N \to  \frac{\beta \eps }{1-\beta}e$ as $k\to\infty.$
Since $\eps$ can be chosen to be arbitrarily close to 0, for all sample paths in $A_N$, we have that:
\begin{align*}
    \limsup_{k\to \infty} \overline{X}_k^N \leq 0.
\end{align*} 

Since the union of the events $A_N$ is the entire sample space, we have:
\begin{align*}
\limsup_{k\to \infty} \overline{X}_k\leq 0.
\end{align*}

From the definition of $\Delta_k$ and the fact that $Y_k \to 0$ a.s., we conclude that:
\begin{align*}
    \limsup_{k\to \infty} \norm{\Delta_k}_\infty= \limsup_{k\to \infty} \norm{V_k - J^*}_\infty \leq 0 a.s,
\end{align*} and hence $V_k \to J^* a.s.$

\section{PROOF OF THEOREM 2}

We define $V_k := \Phi \theta_k$ and write the sequence of iterates $\{V_k\}_{k=0}^\infty$ as follows:
\begin{align*}
    V_{k+1} = (1-\gamma_k)V_k + \gamma_k (H(V_k)+z_k),
\end{align*}
where $H(V_k) = E[\Phi\theta^{\mu_{k+1}}|\scriptF_k]$ and $z_k := \Phi \theta^{\mu_{k+1}}-E[\Phi\theta^{\mu_{k+1}}|\scriptF_k].$

\textit{Proof Outline:}
We can use our assumption in Theorem 2 to show that:
\begin{align*}
    &\norm{H(V_k)-T^{H-1}V_k}_\infty 
    \leq  \kappa\frac{\alpha^{H-1}}{1-\alpha}\norm{TV_k - V_k}_\infty+ \delta_{app},
\end{align*} for some $\kappa$ and $\delta_{app}$, which implies that: 
\begin{align*}
  \norm{H(V_k)-J^*}_\infty  \leq  \underbrace{\Big(\alpha^{H-1}+\kappa\frac{2\alpha^{H-1}}{1-\alpha}\Big)}_{=: \beta}\norm{V_k - J^*}_\infty +\delta_{app}.
\end{align*}

Thus, $H(V_k)$ becomes almost a contraction with an error term, $\delta_{app}$. We can then apply stochastic approximation techniques to show that:
\begin{align*}
    \limsup_{k\to \infty} \norm{V_k - J^*}_\infty \leq  \frac{\delta_{app}}{1-\beta}.
\end{align*}

To see this, suppose that there is no noise and so our iterates do not involve the noise averaging, i.e., 
$$
V_{k+1} = \underbrace{\hat{J}^{\mu_{k+1}}}_{=: H(V_k)},
$$
where $\norm{\hat{J}^{\mu_{k+1}}-J^{\mu_{k+1}}}_\infty\leq \delta.$
Then, we can trace the steps of the above, defining $\kappa$ and $\delta_{app}$ as we did above and we have the following:
\begin{align*}
   &T^{H-1}V_k - \kappa\frac{\alpha^{H-1}}{1-\alpha}\norm{TV_k-V_k}_\infty -\delta_{app} \leq V_{k+1} \leq T^{H-1}V_k + \kappa\frac{\alpha^{H-1}}{1-\alpha}\norm{TV_k-V_k}_\infty +\delta_{app} \\
   &\implies \norm{V_{k+1}-T^{H-1}V_k}_\infty \leq \kappa\frac{\alpha^{H-1}}{1-\alpha}\norm{TV_k-V_k}_\infty +\delta_{app} \\
   &\implies \norm{V_{k+1}-J^*}_\infty -\norm{T^{H-1}V_k-J^*}_\infty \leq \kappa\frac{\alpha^{H-1}}{1-\alpha}\norm{TV_k-V_k}_\infty +\delta_{app} \\ 
   &\implies \norm{V_{k+1}-J^*}_\infty  \leq\norm{T^{H-1}V_k-J^*}_\infty+
   \kappa\frac{\alpha^{H-1}}{1-\alpha}\norm{TV_k-V_k}_\infty +\delta_{app} \\
      &\implies \norm{V_{k+1}-J^*}_\infty  \leq \alpha^{H-1}\norm{V_k-J^*}_\infty+ \kappa\frac{\alpha^{H-1}}{1-\alpha}(1+\alpha)\norm{V_k - J^*}_\infty +\delta_{app} \\
          &\implies \norm{V_{k+1}-J^*}_\infty  \leq  \Big(\alpha^{H-1}+\kappa\frac{\alpha^{H-1}}{1-\alpha}(1+\alpha)\Big)\norm{V_k - J^*}_\infty +\delta_{app} \\
   &\implies \norm{V_k - J^*} \leq \Big(\alpha^{H-1}+\kappa\frac{\alpha^{H-1}}{1-\alpha}(1+\alpha)\Big)^k\norm{V_0-J^*} + \delta_{app}\sum_{i=0}^{k-1}
\Big(\alpha^{H-1}+\kappa\frac{\alpha^{H-1}}{1-\alpha}(1+\alpha)\Big)^i \end{align*}
Taking limits, we get the following:
\begin{align*}
    \limsup_{k \to \infty} \norm{V_k-J^*}_\infty &\leq \frac{\delta_{app}}{1-\alpha^{H-1}-\kappa\frac{2\alpha^{H-1}}{1-\alpha}}\\
    &=\frac{\delta_{app}}{1-\beta}.
\end{align*}

We will now prove our Theorem. We break the proof up into steps.

\noindent\textit{Step 1:} We first obtain an upper bound for $\norm{H(V_k)-T^{H-1}V_k}_\infty $ as follows:
\begin{align*}
    &\norm{H(V_k)-T^{H-1}V_k}_\infty 
    \leq   \kappa\norm{  T^{H-1}V_k -J^{\mu_{k+1}}}_\infty+ \delta_{app} 
\end{align*}

\textit{Proof of Step 1:} We assume the existence of $\kappa$ and $\delta_{app}$ in the statement of Theorem 2.

\noindent\textit{Step 2:} 
\begin{align*}
    \norm{J^{\mu_{k+1}}-T^{H-1}V_k}_\infty \leq \frac{\alpha^{H-1}}{1-\alpha}\norm{TV_k - V_k}_\infty.
\end{align*}

\textit{Proof of Step 2:}
Step 2 is a restatement of Lemma 1, which is proved in Appendix B.

\noindent\textit{Step 3:} 
\begin{align*}
    &\norm{H(V_k)-T^{H-1}V_k}_\infty 
    \leq  (1+\kappa)\frac{\alpha^{H-1}}{1-\alpha}\norm{TV_k - V_k}_\infty+ \delta_{app} 
\end{align*}
\textit{Proof of Step 3:}

We have from Step 1:
\begin{align*}
    \norm{H(V_k)- T^{H-1}V_k}_\infty \leq \kappa \norm{  T^{H-1}V_k -J^{\mu_{k+1}}}_\infty +  \delta_{app}.
\end{align*}

Plugging in the result of Step 2, we have:
\begin{align*}
    \norm{H(V_k)- T^{H-1}V_k}_\infty \leq \kappa\frac{\alpha^{H-1}}{1-\alpha}\norm{TV_k - V_k}_\infty +  \delta_{app}.
\end{align*}

\noindent\textit{Step 4:} 
\begin{align*}
  \norm{H(V_k)-J^*}_\infty  \leq  \underbrace{\Big(\alpha^{H-1}+\kappa\frac{2\alpha^{H-1}}{1-\alpha}\Big)}_{=: \beta}\norm{V_k - J^*}_\infty +\delta_{app}.
\end{align*}
\textit{Proof of Step 4:}
We have
\begin{align*}
   &T^{H-1}V_k - \kappa\frac{\alpha^{H-1}}{1-\alpha}\norm{TV_k-V_k}_\infty e -\delta_{app} e\leq H(V_k) \nonumber\\&\leq T^{H-1}V_k + \kappa\frac{\alpha^{H-1}}{1-\alpha}\norm{TV_k-V_k}_\infty e+\delta_{app} e \\
   &\implies \norm{H(V_k)-T^{H-1}V_k}_\infty \leq \kappa\frac{\alpha^{H-1}}{1-\alpha}\norm{TV_k-V_k}_\infty +\delta_{app} \\
   &\implies \norm{H(V_k)-J^*}_\infty -\norm{T^{H-1}V_k-J^*}_\infty \leq \kappa\frac{\alpha^{H-1}}{1-\alpha}\norm{TV_k-V_k}_\infty \nonumber\\&+\delta_{app} \\ 
   &\implies \norm{H(V_k)-J^*}_\infty  \leq\norm{T^{H-1}V_k-J^*}_\infty+
   \kappa\frac{\alpha^{H-1}}{1-\alpha}\norm{TV_k-V_k}_\infty \nonumber\\&+\delta_{app} \\
      &\implies \norm{H(V_k)-J^*}_\infty  \leq \alpha^{H-1}\norm{V_k-J^*}_\infty
      \nonumber\\&+ \kappa\frac{\alpha^{H-1}}{1-\alpha}(1+\alpha)\norm{V_k - J^*}_\infty +\delta_{app} \\
          &\implies \norm{H(V_k)-J^*}_\infty  \leq  \underbrace{\Big(\alpha^{H-1}+\kappa\frac{2\alpha^{H-1}}{1-\alpha}\Big)}_{=: \beta}\norm{V_k - J^*}_\infty \nonumber\\&+\delta_{app}.
\end{align*}
\noindent\textit{Step 5:} 
\begin{align*}
    \limsup_{k\to \infty} \norm{V_k - J^*}_\infty \leq  \frac{\delta_{app}}{1-\beta}.
\end{align*}
\textit{Proof of Step 5:}

So far, we have the following rewrite of our iterates:
\begin{align*}
    V_{k+1} = (1-\gamma_k) V_k + \gamma_k (H(V_k) + z_k),
\end{align*}
where 
\begin{align*}
    \norm{H(V_k)-J^*}_\infty \leq \beta\norm{V_k-J^*}_\infty + \delta_{app}.
\end{align*}

We define $\Delta_{k}:= V_{k}-J^*$. Using $\Delta_{k},$ the following holds:
\begin{align*}
   \Delta_{k+1} = (1-\gamma_k) \Delta_k + \gamma_k (H(V_k)-J^*+w_k). 
\end{align*}

Letting $Y_{k}$ be a sequence defined recursively as follows:
\begin{align*}
    Y_{k+1} = (1-\gamma_k) Y_k + \gamma_k w_k,
\end{align*} where $Y_0 =0.$
Since $w_k$ is bounded for all $k$, $Y_k \to 0$ a.s.

We now define the following sequence $X_k$ as follows:
\begin{align}
    X_k := \Delta_k - Y_k.\label{eq:Xk}
\end{align}

Thus, 
\begin{align*}
    X_{k+1} = (1-\gamma_k)X_k + \gamma_k (H(V_k)-J^*).
\end{align*}

Taking norms on both sides gives:
\begin{align*}
    \norm{X_{k+1}}_\infty &\leq (1-\gamma_k)\norm{X_k}_\infty + \gamma_k \norm{H(V_k)-H(J^*)}_\infty \\
    &\leq (1-\gamma_k)\norm{X_k}_\infty + \gamma_k \Big[\beta\norm{V_k-J^*}_\infty + \delta_{app}\Big] \\
    &=\leq (1-\gamma_k)\norm{X_k}_\infty + \gamma_k \Big[\beta\norm{\Delta_k}_\infty + \delta_{app}\Big] \\
    &\leq (1-\gamma_k)\norm{X_k}_\infty + \gamma_k \Big[\beta\norm{X_k}_\infty +\beta \norm{Y_k}_\infty+ \delta_{app}\Big].
\end{align*}

Since $Y_k \to 0$ a.s., we conclude there must exist for all $\eps>0$ some $k(\eps)$ such that for all $k>k(\eps):$
$$
\norm{Y_k}_\infty \leq \eps. 
$$
So, for $k>k(\eps),$ the following holds:
\begin{align*}
    \norm{X_{k+1}}_\infty &\leq  (1-\gamma_k)\norm{X_k}_\infty + \gamma_k \Big[\beta\norm{X_k}_\infty +\beta \eps+ \delta_{app}\Big].
\end{align*}
Rearranging terms, we have:
\begin{align*}
    &\norm{X_{k+1}}_\infty \nonumber\\&\leq (1-\gamma_k(1-\beta))\norm{X_k}_\infty + \gamma_k(\beta \eps+\delta_{app}) \\
    &= (1-\underbrace{\gamma_k(1-\beta)}_{=: \gamma_k'})\norm{X_k}_\infty + \gamma_k (1-\beta)\Big[ \frac{\beta \eps+\delta_{app}}{1-\beta}\Big]. 
\end{align*}

Now, consider any positive integer $N$. We define a sequence of random variables $\overline{X}_k^N$ for $k \geq N,$ by setting $\overline{X}_N^N= \norm{X_N}_\infty$ and 
\begin{align*}
    \overline{X}_{k+1}^N = (1-\gamma_k')\overline{X}_k^N + \gamma_k' \Big[ \frac{\beta \eps+\delta_{app}}{1-\beta}\Big]\forall k > N.
\end{align*}

We will carry out a comparison of the sequence $\norm{X_k}_\infty$ with the sequence $\overline{X}_k^N$.
Consider the event that $k(\eps)=N,$ which we denote by $A_N.$ We can use an easy inductive argument to show that for any $N$, for any sample path in $A_N,$ and for all $k\geq N,$ that $\norm{X_k}_\infty \leq \overline{X}_k^N.$
It is evident from the assumption that $\sum_{k=0}^\infty \gamma_k = \infty$ and thus $\sum_{k=0}^\infty \gamma_k' = \infty $ that $\overline{X}_k^N \to  \frac{\beta \eps+\delta_{app}}{1-\beta}$ as $k\to\infty.$
Since $\eps$ can be chosen to be arbitrarily close to 0, for all sample paths in $A_N$, we have that
\begin{align*}
    \limsup_{k\to \infty} \overline{X}_k^N \leq \frac{\delta_{app}}{1-\beta}.
\end{align*}

Since the union of the events $A_N$ is the entire sample space, we have:

\begin{align*}
\limsup_{k\to \infty} \overline{X}_k\leq \frac{\delta_{app}}{1-\beta}.
\end{align*}

From the definition of $\Delta_k$ and the fact that $Y_k \to 0$ a.s., we conclude that 
\begin{align*}
    \limsup_{k\to \infty} \norm{\Delta_k}_\infty= \limsup_{k\to \infty} \norm{V_k - J^*}_\infty \leq  \frac{\delta_{app}}{1-\beta}.
\end{align*}

Furthermore, since $V_k = \Phi \theta_k,$ we have that 
\begin{align*}
   \limsup_{k\to \infty} \norm{\Phi \theta_k- J^*}_\infty \leq  \frac{\delta_{app}}{1-\beta}.
\end{align*}


\noindent\textit{Step 6:} 
\begin{align*}
    \limsup_{k\to \infty} \norm{J^{\mu_k} - J^*}_\infty \leq  \frac{\delta_{app}}{1-\beta}.
\end{align*}
\textit{Proof of Step 6:}
Choose any $\eps>0.$ Then, there exists $k(\eps)$ such that the following holds for all $k>k(\eps)$:
\begin{align}
   \norm{V_k-J^*}_\infty \leq \Delta +\eps. \label{eq:policy 1 TAC NO NOISE}
\end{align}

Using \eqref{eq:policy 1 TAC NO NOISE}, we can see that: 
\begin{align*}
&\norm{V_k -TV_k}_\infty - \norm{TV_k - J^*}_\infty\leq \norm{V_k-TV_k +TV_k -J^*}_\infty  \norm{V_k-J^*}_\infty \leq \Delta +\eps\\
&\implies \norm{V_k - TV_k }_\infty \leq \norm{TV_k - J^*}_\infty+ \Delta +\eps\\
&\implies \norm{V_k - TV_k }_\infty \leq \alpha\norm{V_k - J^*}_\infty+ \Delta +\eps\\
&\implies \norm{V_k - TV_k }_\infty \leq \alpha(\Delta+\eps)+ \Delta +\eps\\
&\implies \norm{V_k - TV_k }_\infty \leq(1+ \alpha) (\Delta +\eps).
\end{align*}

Thus, 
\begin{align}
  \nonumber &- TV_k \leq -V_k + (1+ \alpha) (\Delta +\eps)e \\\nonumber
   \implies& -T^H V_k \leq - T^{H-1}V_k +\alpha^{H-1} (1+ \alpha) (\Delta +\eps)e
   \\\nonumber
   \implies& -T_{\mu_{k+1}} T^{H-1} V_k \leq - T^{H-1}V_k +\alpha^{H-1} (1+ \alpha) (\Delta +\eps)e.
\end{align}

Suppose that we apply the $T_{\mu_{k+1}}$ operator $\ell-1$ times. Then, due to monotonicity and the fact that $T_\mu(J+ce)=T_\mu(J)+\alpha ce,$ for any policy $\mu,$ we have the following:
\begin{align*}
     -T_{\mu_{k+1}}^{\ell} T^{H-1} V_k \leq - T_{\mu_{k+1}}^{\ell-1}T^{H-1}V_k +\alpha^{\ell-1}\alpha^{H-1} (1+ \alpha) (\Delta +\eps)e.
\end{align*}

Using a telescoping sum, we get the following inequality:
\begin{align*}
   - T_{\mu_{k+1}}^j T^{H-1} V_k + T^{H-1}V_k 
    &\leq -\sum_{\ell = 1}^{j} \alpha^{\ell-1}  \alpha^{H-1}(\alpha+1)(\Delta +\eps)e.
\end{align*}

Taking the limit as $j\rightarrow\infty$ on both sides, we have the following:
\begin{align*}
    -J^{\mu_{k+1}} + T^{H-1}V_k \leq -\frac{\alpha^{H-1}(\alpha+1)(\Delta +\eps)}{1-\alpha}e.
\end{align*}

Rearranging terms and subtracting $J^*$ from both sides, we get the following:
\begin{align*}
&-J^{\mu_{k+1}} + T^{H-1}V_k \leq -\frac{\alpha^{H-1}(\alpha+1)(\Delta +\eps)}{1-\alpha}e \\
&\implies  J^*-J^{\mu_{k+1}}\leq J^* -T^{H-1}V_k-\frac{\alpha^{H-1}(\alpha+1)(\Delta +\eps)}{1-\alpha}e 
\end{align*}
Since $J^\mu \leq J^*$ for all policies $\mu$, the above line implies that:
\begin{align*}
\norm{J^*-J^{\mu_{k+1}}}_\infty &\leq \norm{J^* -T^{H-1}V_k}_\infty +\frac{\alpha^{H-1}(\alpha+1)(\Delta +\eps)}{1-\alpha} \\
&\leq \alpha^{H-1}\norm{J^* -V_k}_\infty +\frac{\alpha^{H-1}(\alpha+1)(\Delta +\eps)}{1-\alpha} \\
&\leq \alpha^{H-1}(\Delta+\eps) +\frac{\alpha^{H-1}(\alpha+1)(\Delta +\eps)}{1-\alpha} 
\\   &= \frac{2\alpha^{H-1} (\Delta+\eps)}{1-\alpha}.
\end{align*}

Since the above holds for all $\eps>0,$ we have the following conclusion:
\begin{align*}
  \limsup_{k\to\infty} \norm{ J^{\mu_{k+1}}-J^*}_\infty&\leq  \frac{2\alpha^{H-1}\Delta}{1-\alpha}.
\end{align*}

\section{PROOF OF THEOREM 3 AND EXPLANATION}
Our iterates are:
\begin{align*}
V_{k+1} = (1-\gamma_k)V_k + \gamma_k (H(V_k)+z_k).
\end{align*}

We have 
\begin{align*}
&X_{k+1} = (1-\gamma_k)X_k + \gamma_k (H(V_k))\\
&\implies \norm{X_{k+1}-J^*}_\infty = (1-\gamma_k)\norm{X_{k}-J^*}_\infty + \gamma_k \norm{H(V_k)-J^*}_\infty \\
&\implies \norm{X_{k+1}-J^*}_\infty \leq (1-\gamma_k)\norm{X_{k}-J^*}_\infty + \gamma_k ( (\alpha^{H-1} +\frac{2\kappa\alpha^{H-1}}{1-\alpha})\norm{V_k - J^*}_\infty\\&+ \delta_{app}) \\
&\implies \norm{X_{k+1}-J^*}_\infty \leq (1-\gamma_k(1-\alpha^{H-1}+\frac{2\kappa\alpha^{H-1}}{1-\alpha}))\norm{X_{k}-J^*}_\infty + \gamma_k \delta_{app} \\
&\implies E[\norm{X_{k+1}-J^*}_\infty] \leq (1-\gamma_k(1-\alpha^{H-1}+\frac{2\kappa\alpha^{H-1}}{1-\alpha}))E[\norm{X_{k}-J^*}_\infty] + \gamma_k \delta_{app} \\
&\implies E[\norm{V_{k+1}-J^*}_\infty] \leq (1-\gamma_k(1-\alpha^{H-1}+\frac{2\kappa\alpha^{H-1}}{1-\alpha}))E[\norm{V_{k}-J^*}_\infty] + \gamma_k (E[\norm{Y_{k+1}}_\infty]+E[\norm{Y_{k}}_\infty]+\delta_{app}),
\end{align*}
where the last line follows from using the triangle inequality and the definition in \eqref{eq:Xk}.

Iterating, we have:
$$E[\norm{V_{k}-J^*}_\infty]  \leq \underbrace{\prod_{i=1}^{k-1} a_{i}\norm{V_0-J^*}_\infty}_{\text{initial condition error}} + \underbrace{\delta_{app}\sum_{j=1}^{k-1} \gamma_j \prod_{\ell=j+1}^{k-1} a_\ell}_{\text{error due to function approximation}}+ \underbrace{\sum_{j=1}^{k-1} \gamma_j(E[\norm{Y_{j+1}}_\infty]+E[\norm{Y_{j}}_\infty]) \prod_{\ell=j+1}^{k-1} a_\ell}_{\text{error due to noise}}.$$

We now obtain an upper bound for the $\norm{Y_j}_\infty$ as follows. From the definition of $Y_k$ in Appendix D, we have the following:
\begin{align*}
E(\norm{Y_k+1}^2) \leq (1-\gamma_k)^2 E(\norm{Y_k}^2)+\gamma_k \sigma^2.
\end{align*} Furthermore, since $\norm{Y_0}=0,$ we can iterate over $k$ to get $\sigma_j$ in Section 4.2.
We note that since $\gamma_k$ is square summable and sums to infinity,  $Y_k\to0$ a.s. Additionally, since $\gamma_k$ is square summable and sums to infinity, we have that $\prod_{i=1}^{k-1} a_{i}\to 0.$ 
\section{SECTION 4.3 - PROOFS}
Recall that from the equation in (6), we rewrite our iterates as follows:
\begin{align*}
    \theta^{\mu_{k+1}} &= (\scriptP_{1,k}\Phi)^+\scriptP_{2,k} (T_{\mu_{k+1}}^m T^{H-1}V_k+w_k),
\end{align*}  
and thus, 
\begin{align*}
    \Phi \theta^{\mu_{k+1}} = \underbrace{\Phi  (\scriptP_{1,k}\Phi)^+\scriptP_{2,k}}_{=:\scriptM_k} (T_{\mu_{k+1}}^m T^{H-1}V_k+w_k)
\end{align*}

 We have:
\begin{align*}
    &\norm{H(V_k)-J^{\mu_{k+1}}}_\infty \\
   &= \norm{E[\scriptM_k( T_{\mu_{k+1}}^m T^{H-1}V_k+w_k)-J^{\mu_{k+1}}|\scriptF_k]}_\infty \\
    &= \norm{E[\scriptM_k( T_{\mu_{k+1}}^m T^{H-1}V_k+w_k)-\scriptM_k( J^{\mu_{k+1}}+ w_k)+\scriptM_k( J^{\mu_{k+1}}+ w_k)-J^{\mu_{k+1}}|\scriptF_k]}_\infty \\
    &\leq \norm{E[\scriptM_k( T_{\mu_{k+1}}^m T^{H-1}V_k+w_k)-\scriptM_k( J^{\mu_{k+1}}+ w_k)|\scriptF_k]}_\infty+\norm{E[\scriptM_k( J^{\mu_{k+1}}+ w_k)-J^{\mu_{k+1}}|\scriptF_k]}_\infty \\
    &\leq \norm{E[\scriptM_k( T_{\mu_{k+1}}^m T^{H-1}V_k+w_k)-\scriptM_k( J^{\mu_{k+1}}+ w_k)|\scriptF_k]}_\infty\\&+\underbrace{\sup_{k, \mu_k}\norm{E[\scriptM_k( J^{\mu_{k+1}}+ w_k)-J^{\mu_{k+1}}|\scriptF_k]}_\infty}_{=: \delta_{app}} \\ 
    &= \norm{E[\scriptM_k( T_{\mu_{k+1}}^m T^{H-1}V_k)-\scriptM_k( J^{\mu_{k+1}})|\scriptF_k]}_\infty+ \delta_{app} \\ 
    &= E[\norm{\scriptM_k( T_{\mu_{k+1}}^m T^{H-1}V_k)-\scriptM_k( J^{\mu_{k+1}})}_\infty|\scriptF_k]+ \delta_{app} \\ 
    &\leq  E[\sup_k \norm{\scriptM_k}_\infty \norm{ T_{\mu_{k+1}}^m T^{H-1}V_k- J^{\mu_{k+1}}}_\infty|\scriptF_k]+ \delta_{app} \\ 
    &=  \underbrace{\sup_k \norm{\scriptM_k}_\infty}_{=: \delta_{FV}} \norm{ T_{\mu_{k+1}}^m T^{H-1}V_k -J^{\mu_{k+1}}}_\infty+ \delta_{app} \\ 
    &\leq \alpha^m \delta_{FV} \norm{  T^{H-1}V_k -J^{\mu_{k+1}}}_\infty+ \delta_{app}.
\end{align*}
Using the above, we furthermore have that 
\begin{align*}
    \norm{H(V_k)-T^{H-1}V_k}_\infty \leq  \underbrace{(1+ \alpha^m \delta_{FV})}_{=: \kappa} \norm{  T^{H-1}V_k -J^{\mu_{k+1}}}_\infty+ \delta_{app}.
\end{align*}

\section{SECTION 4.4 - PROOFS}

First, we will show that the gradient descent converges to 
\begin{align}
&{\theta_{k+1}}^* :=\min_\theta \frac{1}{2}\underbrace{\norm{\underbrace{(\scriptP_{1, k}\Phi)}_{=: A_k} \theta - \underbrace{\scriptP_{2, k}\hat{J}^{\mu_{k+1}}}_{=: b_k}}_2^2}_{=: f_k(\theta)}=(\scriptP_{1,k}\Phi)^+ \scriptP_{2,k}(T_{\mu_{k+1}}^m T^{H-1}V_k+w_k). 
\end{align} 

To do so, we will show that 
\begin{align*}
\norm{\theta_{k+1,\eta}-\theta_{k+1}^*}_\infty &\leq  (1-\xi \sigma_{\scriptP_{1,k}\Phi,\max})^\eta\norm{\Phi}_\infty \norm{V_{k,1}}_\infty  \norm{\Sigma_{k,1}^{-1}}_\infty \norm{U_k^\top \scriptP_{2,k}\hat{J}^{\mu_{k+1}}}_\infty,
\end{align*}
where the singular value decomposition of $A_k$ is:
$$
A_k 
= U_k\begin{bmatrix}
\Sigma_{k,1}&
0
\end{bmatrix}\begin{bmatrix}
V_{k,1}^\top \\
V_{k,2}^\top
\end{bmatrix} = U_k \Sigma_{k,1}V_{k,1}^\top.$$
where $U_k$ is a unitary matrix, $\Sigma_{k,1}$ is a rectangular diagonal matrix, and $V_k$ is a unitary matrix.

Note that using the singular value decomposition of $A_k,$ we can rewrite $\theta_{k+1}^*$ as follows:
\begin{align*}
\theta_{k+1}^* = V_{k,1} \Sigma_{k,1}^{-1} U_k^\top b_k.
\end{align*}

The gradient of $f_k(\theta)$ is:
\begin{align*}
    \nabla f_k(\theta) = A_k^\top (A_k\theta-b_k).
\end{align*}

Using gradient descent with step size $\xi> 0,$ our iterates of gradient descent are given by:

\begin{align*}
\theta_{k+1, \ell} &= \theta_{k+1,\ell-1} -\xi \nabla f_k(\theta_{k+1,\ell-1})  \\
&= (I- \xi A_k^\top A_k)\theta_{k+1,\ell-1} + \xi A_k^\top b.
\end{align*}

Hence, 
\begin{align*}
\theta_{k+1,\ell}  
&= \xi \sum_{\ell=0}^{\ell-1} (I- \xi A_k^\top A_k)^\ell A_k^\top b.
\end{align*}
 
From the singular value decomposition of $A_k,$ we have that $$(I - \xi A_k^\top A_k)^\ell = V_k {(I- \xi\Sigma^2_k)}^\ell V_k^\top,$$ we can rewrite $\theta_{k+1,\ell} $ as follows:
\begin{align*}
\theta_{k+1, \ell} 
&= \xi \sum_{\ell=0}^{\ell-1} V_k {(I- \xi\Sigma^2_{k})}^\ell V_k^\top A_k^\top b_k \\
&= \xi \sum_{\ell=0}^{\ell-1} V_k {(I- \xi\Sigma^2_{k}})^\ell \Sigma_k U_k^\top b_k \\
&= \xi \sum_{\ell=0}^{\ell-1} V_k \begin{bmatrix}
(I-\xi \Sigma_{k,1}^2)^\ell  &  0\\
0 & 0
\end{bmatrix} \begin{bmatrix}
\Sigma_{k,1} \\
0
\end{bmatrix} U_k^\top b_k \\
&= \xi \sum_{\ell=0}^{\ell-1} V_{k,1} {(I- \xi \Sigma^2_{k,1})}^\ell \Sigma_{k,1} U_k^\top b_k.
\end{align*}

Since 
$$
\Sigma_{k,1}^{-1}  = \xi(I-I+\xi \Sigma_{k,1}^2)^{-1} \Sigma_{k,1} = \xi \sum_{\ell=0}^\infty (I-\xi \Sigma_{k,1}^2)^\ell \Sigma_{k,1},
$$
we further rewrite $\theta_{k+1}^*$ as follows:
\begin{align*}
\theta_{k+1}^*= \sum_{\ell=0}^\infty \xi V_{k,1} {(I- \xi\Sigma^2_{k,1} )}^\ell \Sigma_{k,1} U_k^\top b_k.
\end{align*}

We now compute 
\begin{align*}
&\norm{\theta^{\mu_{k+1}} -\theta_{k+1}^*}_\infty\\ 
&\leq\norm{\theta_{k+1,\eta} -\theta_{k+1}^*}_\infty\\ 
&=\norm{\xi \sum_{\ell=0}^{\eta} V_{k,1} {(I- \xi\Sigma^2_{k,1})}^\ell \Sigma_{k,1} U_k^\top b_k-\xi \sum_{\ell=0}^\infty V_{k,1} {(I- \xi\Sigma^2_{k,1})}^\ell \Sigma_{k,1} U_k^\top b_k}_\infty\\
&= \norm{\xi \sum_{\ell=\eta}^\infty V_{k,1} {(I- \xi\Sigma^2_{k,1})}^\ell \Sigma_{k,1} U_k^\top b_k}_\infty\\
&\leq \norm{V_{k,1}}_\infty  \norm{ \xi \sum_{\ell=\eta}^\infty {(I- \xi\Sigma^2_{k,1})}^\ell \Sigma_{k,1} U_k^\top b_k}_\infty\\
&= \norm{V_{k,1}}_\infty  \norm{ \xi (I- \xi\Sigma^2_{k,1})^\eta \sum_{\ell=0}^\infty {(I- \xi\Sigma^2_{k,1})}^\ell \Sigma_{k,1} U_k^\top b_k}_\infty\\
&= \norm{V_{k,1}}_\infty \norm{(I- \xi\Sigma^2_{k,1})^\eta}_\infty \norm{ \xi  \sum_{\ell=0}^\infty {(I- \xi\Sigma^2_{k,1})}^\ell \Sigma_{k,1} U_k^\top b_k}_\infty\\
&\leq \norm{V_{k,1}}_\infty \norm{I- \xi\Sigma^2_{k,1}}_\infty^\eta \norm{ \xi  \sum_{\ell=0}^\infty {(I- \xi\Sigma^2_{k,1})}^\ell \Sigma_{k,1} U_k^\top b_k}_\infty\\
&= \norm{V_{k,1}}_\infty \norm{I- \xi\Sigma^2_{k,1}}_\infty^\eta \norm{ \Sigma_{k,1}^{-1} U_k^\top b_k}_\infty\\
&\leq  (1-\xi \sigma_{\scriptP_{1,k}\Phi,\max})^\eta \norm{V_{k,1}}_\infty  \norm{\Sigma_{k,1}^{-1}}_\infty\norm{U_k^\top b_k}_\infty \\
&\leq  (1-\xi \sigma_{\scriptP_{1,k}\Phi,\max})^\eta \norm{V_{k,1}}_\infty  \norm{\Sigma_{k,1}^{-1}}_\infty \norm{U_k^\top \scriptP_{2,k}\hat{J}^{\mu_{k+1}}}_\infty,
\end{align*}
where $\sigma_{\scriptP_{1,k}\Phi,\max}$ is the largest singular value squared of $\scriptP_{1,k}\Phi.$

Note that the above implies that in order to obtain convergence of $\theta_{k+1,\eta}$ to $\theta_{k+1}^*$ as a function of $\eta$, we must have that $0<\xi\sigma_{\scriptP_{1,k}\Phi,\max} <1.$

Thus, we have:
\begin{align*}
    \norm{H(V_k)-\Phi \theta_{k+1}^*}_\infty \leq (1-\xi \sigma_{\scriptP_{1,k}\Phi,\max})^\eta \norm{\Phi}_\infty \norm{V_{k,1}}_\infty  \norm{\Sigma_{k,1}^{-1}}_\infty \norm{U_k^\top \scriptP_{2,k}\hat{J}^{\mu_{k+1}}}_\infty.
\end{align*}

Defining $\eps :=(1-\xi \sigma_{\scriptP_{1,k}\Phi,\max})^\eta \norm{\Phi}_\infty \norm{V_{k,1}}_\infty  \norm{\Sigma_{k,1}^{-1}}_\infty \norm{U_k^\top \scriptP_{2,k}\hat{J}^{\mu_{k+1}}}_\infty$ and using $\scriptM_k$ and $\delta_{app}$ as defined in Appendix E, we obtain:
\begin{align*}
    &\norm{H(V_k)-J^{\mu_{k+1}}}_\infty \\
   &= \norm{E[\scriptM_k( T_{\mu_{k+1}}^m T^{H-1}V_k+w_k)-J^{\mu_{k+1}}|\scriptF_k]}_\infty +\eps\\
    &= \norm{E[\scriptM_k( T_{\mu_{k+1}}^m T^{H-1}V_k+w_k)-\scriptM_k( J^{\mu_{k+1}}+ w_k)+\scriptM_k( J^{\mu_{k+1}}+ w_k)-J^{\mu_{k+1}}|\scriptF_k]}_\infty +\eps\\
    &\leq \norm{E[\scriptM_k( T_{\mu_{k+1}}^m T^{H-1}V_k+w_k)-\scriptM_k( J^{\mu_{k+1}}+ w_k)|\scriptF_k]}_\infty+\norm{E[\scriptM_k( J^{\mu_{k+1}}+ w_k)-J^{\mu_{k+1}}|\scriptF_k]}_\infty +\eps\\
    &\leq \norm{E[\scriptM_k( T_{\mu_{k+1}}^m T^{H-1}V_k+w_k)-\scriptM_k( J^{\mu_{k+1}}+ w_k)|\scriptF_k]}_\infty\\&+\underbrace{\sup_{k, \mu_k}\norm{E[\scriptM_k( J^{\mu_{k+1}}+ w_k)-J^{\mu_{k+1}}|\scriptF_k]}_\infty}_{=: \delta_{app}} +\eps\\ 
    &= \norm{E[\scriptM_k( T_{\mu_{k+1}}^m T^{H-1}V_k)-\scriptM_k( J^{\mu_{k+1}})|\scriptF_k]}_\infty+ \delta_{app} +\eps\\ 
    &= E[\norm{\scriptM_k( T_{\mu_{k+1}}^m T^{H-1}V_k)-\scriptM_k( J^{\mu_{k+1}})}_\infty|\scriptF_k]+ \delta_{app}+\eps \\ 
    &\leq  E[\sup_k \norm{\scriptM_k}_\infty \norm{ T_{\mu_{k+1}}^m T^{H-1}V_k J^{\mu_{k+1}}}_\infty|\scriptF_k]+ \delta_{app} +\eps\\ 
    &=  \underbrace{\sup_k \norm{\scriptM_k}_\infty}_{=: \delta_{FV}} \norm{ T_{\mu_{k+1}}^m T^{H-1}V_k J^{\mu_{k+1}}}_\infty+ \delta_{app} +\eps\\ 
    &\leq  \alpha^m \delta_{FV} \norm{  T^{H-1}V_k -J^{\mu_{k+1}}}_\infty+ \delta_{app}+\eps.
\end{align*}

Using the above, we furthermore have that 
\begin{align*}
   \norm{H(V_k)-T^{H-1}V_k}_\infty &\leq  \underbrace{(1+ \alpha^m \delta_{FV})}_{=: \kappa} \norm{  T^{H-1}V_k -J^{\mu_{k+1}}}_\infty+ \delta_{app}\\&+(1-\xi \sigma_{\scriptP_{1,k}\Phi,\max})^\eta \norm{\Phi}_\infty \norm{V_{k,1}}_\infty  \norm{\Sigma_{k,1}^{-1}}_\infty \norm{U_k^\top \scriptP_{2,k}\hat{J}^{\mu_{k+1}}}_\infty.
\end{align*}
The new $\kappa$ and $\delta_{app}$ are apparent from the above.
\section{SECTION 4.5 - PROOFS}
\begin{align*}
(\pi_{\mu_{k+1},\min})^2E[\norm{\Phi{\theta^{\mu_{k+1}}}^*-\Phi \theta^{\mu_{k+1}}}_\infty^2]\leq (\pi_{\mu_{k+1},\min})^2 E[\norm{\Phi{\theta^{\mu_{k+1}}}^*-\Phi \theta^{\mu_{k+1}}}_2^2]\leq E[\norm{\Phi{\theta^{\mu_{k+1}}}^*-\Phi \theta^{\mu_{k+1}}}_D^2]\leq \delta_{T,\mu_{k+1}}.
\end{align*}

Using Jensen's inequality, we have:
\begin{align*}
&(\pi_{\mu_{k+1},\min})^2 (E[\norm{\Phi{\theta^{\mu_{k+1}}}^*-\Phi \theta^{\mu_{k+1}}}_\infty])^2 \leq (\pi_{\mu_{k+1},\min})^2E[\norm{\Phi{\theta^{\mu_{k+1}}}^*-\Phi \theta^{\mu_{k+1}}}_\infty^2]\leq\delta_{T,\mu_{k+1}} \\
\implies  &\pi_{\mu_{k+1},\min} E[\norm{\Phi{\theta^{\mu_{k+1}}}^*-\Phi \theta^{\mu_{k+1}}}_\infty] \leq \sqrt{\delta_{T,\mu_{k+1}}}\\
\implies & E[\norm{\Phi{\theta^{\mu_{k+1}}}^*-\Phi \theta^{\mu_{k+1}}}_\infty] \leq \frac{\sqrt{\delta_{T,\mu_{k+1}}}}{\pi_{\mu_{k+1},\min}} \\
\implies & E[\norm{\Phi{\theta^{\mu_{k+1}}}^*-J^{\mu_{k+1}}+J^{\mu_{k+1}}-\Phi \theta^{\mu_{k+1}}}_\infty] \leq \frac{\sqrt{\delta_{T,\mu_{k+1}}}}{\pi_{\mu_{k+1},\min}} \\
\implies & E[\norm{J^{\mu_{k+1}}-\Phi \theta^{\mu_{k+1}}}_\infty] \leq \sup_{\mu_{k+1}} \norm{\Phi{\theta^{\mu_{k+1}}}^*-J^{\mu_{k+1}}}_\infty +\frac{\sqrt{\delta_{T,\mu_{k+1}}}}{\pi_{\mu_{k+1},\min}} ,
\end{align*} where the last inequality follows from applying the reverse triangle inequality and then taking the supremum over all policies $\mu_{k+1}$.

Finally, we use Jensen's inequality again to obtain the following:
\begin{align*}
\norm{E[\Phi \theta^{\mu_{k+1}}]-J^{\mu_{k+1}}}_\infty = 
\norm{E[J^{\mu_{k+1}}-\Phi \theta^{\mu_{k+1}}]}_\infty \leq \sup_{\mu_{k+1}} \norm{\Phi{\theta^{\mu_{k+1}}}^*-J^{\mu_{k+1}}}_\infty +\frac{\sqrt{\delta_{T,\mu_{k+1}}}}{\pi_{\mu_{k+1},\min}} .
\end{align*}

Thus, we can combine the $\delta_{T, \mu_{k+1}}$  in (Bhandari et al., 2018) with the above terms to obtain a $\delta_{app}$ and our calculations in Section 4.5 give $\kappa=1$.  
\section{CONNECTION OF MONTE CARLO ES TO PRACTICE}

We make several remarks regarding the connection of Monte Carlo ES to practice. While AlphaZero \cite{silver2017mastering} uses techniques such as function approximation and lookahead through planning algorithms in the form of Monte Carlo Tree Search (MCTS), Monte Carlo ES is nonetheless a Monte Carlo algorithm since it uses full trajectories and their returns to estimate loss functions. Additionally, the AlphaZero algorithm uses returns from all states visited by the trajectories to make updates instead of  only the first state.  

\end{document}